\documentclass[11pt]{article}

\usepackage[preprint]{acl}

\usepackage{times}
\usepackage{latexsym}

\usepackage[T1]{fontenc}

\usepackage[utf8]{inputenc}

\usepackage{microtype}

\usepackage{inconsolata}

\usepackage{graphicx}

\usepackage{enumitem}
\usepackage{booktabs}
\usepackage{lipsum}
\usepackage{amsmath}
\usepackage{placeins}
\usepackage{array}

\newcommand{\sys}{\mbox{\sc Theorizer}}

\usepackage[table]{xcolor}
\usepackage{pgf}

\definecolor{ValBlue}{RGB}{49,130,189}
\definecolor{PosGreen}{RGB}{49,163,84}
\definecolor{NegRed}{RGB}{222,45,38}
\definecolor{ValPurple}{RGB}{177,27,232}
\definecolor{ValPink}{RGB}{240,82,156}
\definecolor{ValTeal}{RGB}{15,203,140}

\definecolor{D3Blue}{RGB}{31,119,180}
\definecolor{Purple}{RGB}{129,31,180}
\definecolor{D3Green}{RGB}{44,160,44}
\definecolor{D3Red}{RGB}{214,39,40}

\newcommand{\ValMin}{1}        %
\newcommand{\ValMax}{10}
\newcommand{\DeltaMaxAbs}{2}   %

\newcommand{\val}[1]{%
  \pgfmathsetmacro{\xval}{min(\ValMax, max(\ValMin, #1))}%
  \pgfmathsetmacro{\tval}{(\xval - \ValMin) / (\ValMax - \ValMin)}%
  \pgfmathsetmacro{\pval}{int(\tval * 65)}%
  \edef\temp{\noexpand\cellcolor{D3Blue!\pval!white}}%
  \temp#1%
}
\newcommand{\valpur}[1]{%
  \pgfmathsetmacro{\xval}{min(\ValMax, max(\ValMin, #1))}%
  \pgfmathsetmacro{\tval}{(\xval - \ValMin) / (\ValMax - \ValMin)}%
  \pgfmathsetmacro{\pval}{int(\tval * 65)}%
  \edef\temp{\noexpand\cellcolor{Purple!\pval!white}}%
  \temp#1%
}

\newcommand{\valpu}[1]{%
  \pgfmathsetmacro{\xval}{min(1, max(0, #1))}%
  \pgfmathsetmacro{\tval}{\xval}%
  \pgfmathsetmacro{\pval}{int(\tval * 65)}%
  \edef\temp{\noexpand\cellcolor{Purple!\pval!white}}%
  \temp#1%
}
\newcommand{\valpk}[1]{%
  \pgfmathsetmacro{\xval}{min(1, max(0, #1))}%
  \pgfmathsetmacro{\tval}{\xval}%
  \pgfmathsetmacro{\pval}{int(\tval * 65)}%
  \edef\temp{\noexpand\cellcolor{D3Blue!\pval!white}}%
  \temp#1%
}
\newcommand{\valng}[1]{%
  \pgfmathsetmacro{\xval}{min(1, max(0, #1))}%
  \pgfmathsetmacro{\tval}{\xval}%
  \pgfmathsetmacro{\pval}{int(\tval * 65)}%
  \edef\temp{\noexpand\cellcolor{D3Green!\pval!white}}%
  \temp#1%
}

\newcommand{\vald}[1]{%
  \pgfmathsetmacro{\dval}{#1}%
  \pgfmathsetmacro{\dcval}{min(\DeltaMaxAbs, max(-\DeltaMaxAbs, \dval))}%
  \pgfmathsetmacro{\absval}{abs(\dcval)}%
  \pgfmathsetmacro{\pval}{int((\absval / \DeltaMaxAbs) * 65)}%
  \pgfmathparse{\dcval > 0.01 ? 1 : 0}%
  \let\ispos\pgfmathresult%
  \pgfmathparse{\dcval < -0.01 ? 1 : 0}%
  \let\isneg\pgfmathresult%
  \ifdim\ispos pt=1pt%
    \edef\temp{\noexpand\cellcolor{D3Green!\pval!white}}%
    \temp%
  \else%
    \ifdim\isneg pt=1pt%
      \edef\temp{\noexpand\cellcolor{D3Red!\pval!white}}%
      \temp%
    \fi%
  \fi%
  #1%
}

\newcommand{\DeltaMaxAbsPc}{100}

\newcommand{\valpc}[1]{%
  \pgfmathsetmacro{\dval}{#1}%
  \pgfmathsetmacro{\dcval}{min(\DeltaMaxAbsPc, max(-\DeltaMaxAbsPc, \dval))}%
  \pgfmathsetmacro{\absval}{abs(\dcval)}%
  \pgfmathsetmacro{\pval}{int((\absval / \DeltaMaxAbsPc) * 65)}%
  \pgfmathparse{\dcval > 0.5 ? 1 : 0}%
  \let\ispos\pgfmathresult%
  \pgfmathparse{\dcval < -0.5 ? 1 : 0}%
  \let\isneg\pgfmathresult%
  \ifdim\ispos pt=1pt%
    \edef\temp{\noexpand\cellcolor{D3Green!\pval!white}}%
    \temp%
  \else%
    \ifdim\isneg pt=1pt%
      \edef\temp{\noexpand\cellcolor{D3Red!\pval!white}}%
      \temp%
    \fi%
  \fi%
  #1\%%
}

\title{Generating Literature-Driven Scientific Theories at Scale}

\author{
 \textbf{Peter Jansen\textsuperscript{1,2}},
 \textbf{Peter Clark\textsuperscript{1}}
 \textbf{Doug Downey\textsuperscript{1}},
 \textbf{Daniel S. Weld\textsuperscript{1,3}}
\\
 \textsuperscript{1}Allen Institute for Artificial Intelligence
 \textsuperscript{2}University of Arizona\\
 \textsuperscript{3}University of Washington\\
 \small{
   \texttt{peterj@allenai.org}
 }
}

\begin{document}
\maketitle
\begin{abstract}
Contemporary automated scientific discovery has focused on agents for generating scientific experiments, while systems that perform higher-level scientific activities such as theory building remain underexplored.
In this work, we formulate the problem of synthesizing theories consisting of qualitative and quantitative laws from large corpora of scientific literature.
We study theory generation at scale, using 13.7k source papers to synthesize 2.9k theories, examining how generation using literature-grounding versus parametric knowledge, and accuracy-focused versus novelty-focused generation objectives change theory properties.
Our experiments show that, compared to using parametric LLM memory for generation, our literature-supported method creates theories that are significantly better at both matching existing evidence and
at predicting future results from 4.6k subsequently-written papers.\footnote{\url{https://github.com/allenai/asta-theorizer}} 
\end{abstract}

\section{Introduction}

Automated scientific discovery systems aim to generate novel scientific insights by leveraging data, experiments, and prior literature \cite[e.g.][]{Thilakaratne2019ASRA,pmlr-v235-majumder24a,Tobias2025AutonomousL}. 
Recent advances in language model agents have enabled the integration of these paradigms, leading to a proliferation of experiment-driven discovery systems across domains including computer science, biology, and astronomy \cite[e.g.][]{Lu2024TheAS,jansen-etal-2025-codescientist,schmidgall-etal-2025-agent,Ghafarollahi2024SciAgentsAS,Moss2025TheAC}.

While experiment-driven discovery is central to scientific progress, individual experiments are typically conducted in service of higher-level scientific activities such as theory formation and synthesis \cite{Kuhn1962}. 
Theories consolidate large bodies of empirical results into compact, generalizable statements that support explanation and prediction. For example, Kepler's laws compressed centuries of astronomical observations into a set of laws that describe planetary motion. 
This raises a fundamental question: \emph{can AI systems, built on language models, synthesize scientific theories by reading the research literature?}

\begin{figure*}[t]
  \centering
  \includegraphics[scale=1.03]{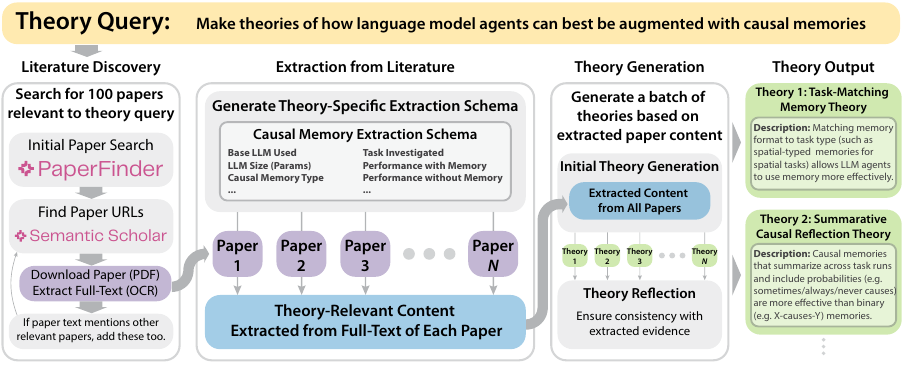}
  \vspace{-2mm}
  \caption{\footnotesize An overview of synthesizing theories from scientific literature with \textsc{Theorizer}.  A user-provided theory query guides a search for scientific papers, then theory-relevant knowledge is extracted from each paper.  That knowledge is provided to a language model which generates and refines a set of theories. Full example theories are large and provided in the \textsc{Appendix}.}
  \label{fig:system_overview}
\end{figure*}

In this work, we first define the problem of literature-based theory synthesis --- generating sets of qualitative and quantitative laws that generalize scientific knowledge in a subect area. We then specify five desirable criteria for these generated theories: specificity, empirical support, predictive accuracy, novelty, and plausibility. We introduce a novel system, \sys, that reads  tens of thousands of papers spanning diverse topics in artificial intelligence and natural language processing to generate numerous candidate theories. We explore several variants of \sys --- a RAG-style, literature-supported method and a simpler  LLM baseline. For each of these methods, we consider two variants --- one that optimizes accuracy and another that aims to increase novelty. 

Evaluating thousands of theories via new experiments is infeasible at scale.  To address this, we develop several evaluation methods, including  a backtesting paradigm \cite{YETISGENYILDIZ2009633} where theories are generated using a fixed knowledge cutoff, and their predictions are evaluated against experimental results reported in subsequently published literature.
Experiments show that, while our literature-supported method is almost 7x more expensive than a simple LLM approach, it yields significantly more accurate and predictive theories - as measured both by using an LLM judge on theory-law statements, and by comparing the generated theories to 4,554 subsequently-written papers in our backtesting paradigm. 
We further analyze how theory quality varies across the generation conditions, comparing generation using literature-supported versus purely parametric knowledge, and accuracy-focused versus novelty-focused objectives. In summary, we make the following  contributions:

\begin{enumerate}
    \item We formulate the problem of literature-based scientific theory synthesis.
    \item We present the \sys\ system, which has created 2,856 theories by synthesizing evidence from 13,744 scientific papers.
    \item Using \sys, we systematically compare theory generation using literature-supported versus parametric knowledge, and accuracy-focused versus novelty-focused objectives. We  analyze how these conditions affect theories'  specificity, empirical support, predictive accuracy, novelty, and plausibility. To measure predictive accuracy, we use a backtesting paradigm, evaluating how well each generated law matches results from 4,554 papers that were published subsequent to \sys's knowledge-cutoff.
\item Our experiments show that our literature-supported theories  are significantly more accurate matching {\em existing} evidence and also at predicting {\em future} results.%
\end{enumerate}

\section{Related Work}

Scientific theories are typically described as sets of qualitative and quantitative laws that predict empirical observations given initial experimental conditions, while offering some explanatory mechanism for these observations \cite{dvzeroski2007computational}. 
Prior work in automated scientific discovery has largely focused on inducing either quantitative or qualitative laws, typically from structured data rather than from the research literature. 
Our work addresses recent calls for integrated discovery systems that synthesize multiple forms of scientific knowledge into higher-level theories \cite{Langley2024IntegratedSF}.

\paragraph{Quantitative Law Induction.}
Quantitative laws make specific numerical commitments, such as \textit{``force is equal to mass times acceleration''}.
A substantial body of work studies the discovery of quantitative laws, often framed as symbolic search or equation discovery problems predating language models \cite[e.g.][]{langleybook}. 
More recent systems combine neural and symbolic methods to rediscover known physical laws, such as Kepler’s third law, from data \cite{Cornelio2023CombiningDA}. 
Language models have also been incorporated into search-based frameworks for quantitative discovery, including evolutionary approaches such as FunSearch \cite{RomeraParedes2023MathematicalDFA} and LLM-SR \cite{Shojaee2024LLMSRSEA}, as well as probabilistic modeling systems like BoxLM \cite{pmlr-v235-li24v}. 
Several benchmarks evaluate data-driven scientific discovery \cite{majumder2025discoverybench,Cranmer2023InterpretableMLA,gu-etal-2024-blade}. 
In contrast to this line of work, we focus on synthesizing theories from experimental evidence reported across up to one hundred papers in the scientific literature per theory, rather than inducing equations directly from raw data.

\paragraph{Qualitative Law Induction.}
Qualitative laws capture relational or directional regularities without precise numerical commitments, such as \textit{``acids and alkakis react to produce salts''}, and play a central role in scientific reasoning \cite{Forbus1984QualitativePT,Forbus2019QualitativeRepresentations}. 
Prior work has extracted qualitative relationships from text to support tasks such as information extraction, causal modeling, and rule induction \cite[e.g.][]{Dagdelen2024StructuredIEA,friedman2022unstructuredtextcausalknowledge,yang-etal-2024-language}. 
Historical systems for qualitative scientific discovery typically relied on logical or symbolic induction \cite{Langley1985DiscoveringQE,King1996StructureactivityRD}. 
More recently, language models have been used to induce qualitative relations from parametric knowledge alone \cite{bosselut-etal-2019-comet,Bouraoui2019InducingRKA}. 
Despite this progress, inducing qualitative scientific laws from empirical evidence remains relatively under-explored, and in this work we synthesize empirically-grounded qualitative laws by explicitly aggregating experimental evidence across  collections of scientific papers.

\section{Problem Formulation}

We formulate the task of \emph{theory synthesis from scientific literature} as follows.
The synthesis process takes as input a user-specified \emph{theory query} $(Q)$ and a corpus of scientific papers $(C)$, and produces a set of theories $(T)$: $(Q, C) \rightarrow T$.
Each theory $t_i \in T$ consists of a set of $j$ structured \textsc{<law, scope, evidence>} tuples,
$\langle l_j, s_j, e_j \rangle$.
A \emph{law} $(l_j)$ expresses a qualitative or quantitative relationship, represented either as a mathematical expression or a natural-language statement (e.g., \emph{$X$ causes $Y$}, \emph{$A$ increases $B$}, \emph{$X = Y / Z$}).
The \emph{scope} $(s_j)$ specifies the conditions under which the law is expected to hold, including domain constraints or known exceptions (e.g., \emph{applies only for small $R$}, \emph{does not apply when $P$ is present}).
The associated \emph{evidence} $(e_j)$ lists empirical or observational support for the law, typically drawn from the input corpus.

In addition to its constituent laws, each theory $t_i$ has a natural-language \emph{theory name} $(n_i)$ and a high-level \emph{theory description} $(d_i)$, which together provide contextual grounding and situate the theory within the broader scientific literature.

\subsection{Theory Desiderata}
\label{sec:theory-desiderata}
A high-quality scientific theory and its component laws should include these desirable characteristics:
\begin{enumerate}[noitemsep, topsep=0pt]
   \item \textit{Specificity:} Laws should make specific, testable claims and predictions.    \item \textit{Empirical Support:} These claims should be consistent with existing empirical evidence.
    \item \textit{Predictive Accuracy:} These claims should be supported when evaluated against future or held-out experimental results.
    \item \textit{Novelty:} Laws should introduce insights not explicitly stated in prior work.
    \item \textit{Plausibility:} Laws should have a plausible scientific rationale or mechanistic explanation.
\end{enumerate}

\noindent
We operationalize approximate, automated measures of these desiderata in
Section~\ref{sec:experiments}.

\section{Theory Generation}
\label{sec:theory-generation}

Figure~\ref{fig:system_overview} provides an overview of \textsc{Theorizer}, a system for inducing scientific theories from the research literature given a user-specified \textit{theory query}.  The system discovers relevant papers, extracts structured evidence, and uses this evidence to synthesize structured theories.  A system description is below, with additional system details and hyperparameters provided in the \textsc{Appendix}, and source code provided in the source repository.

\subsection{Literature Discovery}

Given a theory query $Q$, \textsc{Theorizer} builds corpus $C$ by retrieving up to $K$ relevant research papers to serve as evidence. The theory query is first reformulated into a literature search query and submitted to \textsc{Ai2 PaperFinder}.\footnote{\url{https://github.com/allenai/asta-paper-finder}}  For each candidate paper, \textsc{Semantic Scholar} \cite{ammar-etal-2018-construction,Kinney2023TheSS} is used to identify links to open-access PDFs, which are downloaded and converted to full text using an OCR-based extraction pipeline. If fewer than $K$ papers are obtained in the initial search, the system expands the candidate set by searching the full text of retrieved papers for additional references, ranking them by relevance, and backfilling the corpus with the highest-scoring papers. For the experiments reported here, $K$ is set at 100 papers to balance context window token limits and extraction time.

\begin{table*}[t!]
\centering
\footnotesize
\newcolumntype{C}{>{\centering\arraybackslash}p{1.1cm}}
\begin{tabular}{lcCCCcCCC}
\toprule
~ & ~ & \multicolumn{3}{c}{\textbf{Accuracy-Focused Generation}} & ~ & \multicolumn{3}{c}{\textbf{Novelty-Focused Generation}}    \\
\textbf{Dimension}  & ~ & \textbf{Param.} & \textbf{Lit.} & $\Delta$ & ~ & \textbf{Param.} & \textbf{Lit.} & $\Delta$ \\
\midrule
\rowcolor[HTML]{E0E0E0}
\multicolumn{9}{l}{\textsc{LLM-as-a-Judge}}\\
\midrule
\textsc{~~Specificity}                   &   ~ & \valpur{5.3} & \val{6.5} & \valpc{23} $^\ddagger$ &  ~ & \valpur{7.9} & \val{8.0} & \valpc{1} $^\dagger$ \\
\textsc{~~Empirical Support}             &   ~ & \valpur{3.9} & \val{5.8} & \valpc{49} $^\ddagger$ &  ~ & \valpur{2.0} & \val{3.5} & \valpc{75} $^\ddagger$ \\
\rowcolor[HTML]{F0F0F0}
\textsc{~~Predictive Accuracy} & ~ & \multicolumn{7}{c}{\text{Evaluated in Section~\ref{sec:predictive-accuracy}}} \\
\textsc{~~Novelty}                       &   ~ & \valpur{5.8} & \val{5.8} & \valpc{0} ~~ &  ~ & \valpur{6.3} & 6.1 & \valpc{-3} ~~~ \\
\textsc{~~Plausibility}                  &   ~ & \valpur{7.1} & \val{7.9} & \valpc{11} $^\ddagger$ &  ~ & \valpur{5.0} & \val{6.3} & \valpc{26} $^\ddagger$ \\
\midrule
\textsc{Number of Samples (Laws)}        &   ~ & 1187 & 927 & -- &  ~ & 799 & 790 & -- \\
\bottomrule
\end{tabular}
\vspace{-2mm}
\caption{\footnotesize Average \textsc{LLM-as-a-judge} scores comparing parametric-generated theories (\textit{Param.}) with literature-supported theories (\textit{Lit.}) in two conditions: \textit{accuracy-focused} and \textit{novelty-focused} theory generation objectives. \textsc{LLM-as-a-judge} scores rate common dimensions of ideation, on a \textsc{1-10} Likert scale (1 signifies low, 10 signifies high). Base judge model is \textsc{Claude Sonnet 4.5}. $\Delta$ represents the relative difference between the parametric-only and literature-supported score, with positive signifying the literature-grounded score is higher. Inferential statistical testing using one-sided non-parametric bootstrap resampling (N=10,000 resamples); significant at $p < 0.01 (\ddagger)$, $p < 0.05 (\dagger)$.}
\label{tab:llm-as-a-judge}
\end{table*}

\subsection{Evidence Extraction}

To gather evidence to build each theory, the system generates an extraction schema tailored to the theory query, and uses this to extract theory-relevant information from each of the $K$ papers. The schema specifies the entities, variables, and empirical results required to support theory induction.  For example, for queries concerning memory-augmented language models, the schema might include the task type examined (e.g. question answering), specific memory mechanisms used in experiments (e.g. causal memory, spatial memory, etc.), evaluation setting, and performance comparisons (such as task performance with and without using the memory). An inexpensive extraction model is used to populate the schema with document-specific values from each paper.  The output of this stage is a collection of structured, \textsc{JSON}-formatted extraction records, one per paper, which serve as input for theory synthesis.

\subsection{Theory Synthesis}

The theory synthesis stage aggregates extracted evidence across papers and induces candidate theories grounded in that  evidence using a language model prompt.  %
After generation, the initial batch of theories is refined using a self-reflection step \cite{madaan2023selfrefine}, which is applied independently to each theory to improve internal consistency, evidence attribution, and specificity.

\section{Experiments}
\label{sec:experiments}
We investigate two research questions in literature-grounded theory synthesis.  First, how does having access to \textit{literature} during the theory generation process affect theory quality over using \textit{parametric knowledge} alone?  Second, how does steering theory generation toward either \textit{accuracy-focused} or \textit{novelty-focused} generation objectives change the characteristics of model-generated theories.

\begin{figure*}[t!]
  \centering
  \includegraphics[scale=1.50]{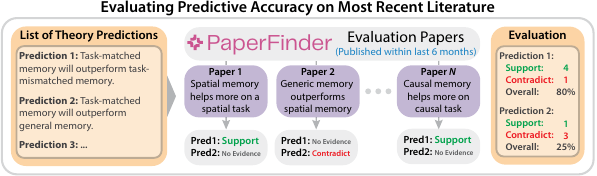}
  \vspace{-2mm}
  \caption{\footnotesize An overview of the \textit{predictive accuracy} evaluation procedure.  For each generated theory law, a language model is used to generate a detailed list of predictions. \textsc{PaperFinder} is used to find papers that may speak to those predictions, and each paper is rated as \textit{supporting}, \textit{contradicting}, or having \textit{no evidence} for each prediction.  This evidence is tallied across papers to arrive at final estimates of predictive precision and recall for a given law.}
  \label{fig:predictive-accuracy-overview}
\end{figure*}

\begin{table*}[t!]
\centering
\footnotesize
\newcolumntype{C}{>{\centering\arraybackslash}p{0.8cm}}
\begin{tabular}{lcCCCcCCC}
\toprule
~ & ~ & \multicolumn{3}{c}{\textbf{Accuracy-Focused}} & ~ & \multicolumn{3}{c}{\textbf{Novelty-Focused}}    \\
\textbf{Dimension}  & ~ & \textbf{Param.} & \textbf{Lit.} & $\Delta$ & ~ & \textbf{Param.} & \textbf{Lit.} & $\Delta$ \\
\midrule
\rowcolor[HTML]{E0E0E0}
\multicolumn{9}{l}{\textsc{Predictive Accuracy (Grounded in Literature)}}\\
\midrule
\textsc{~~Precision: Supporting Evidence}              &   ~ & \valpu{0.88} & \valpk{0.90} & \valpc{2}$^{\dagger}$ &  ~ & \valpu{0.34} & \valpk{0.61} & \valpc{76}$^{\ddagger}$ \\
\textsc{~~Recall: Laws with Some Evidence}      &   ~ & \valpu{0.45} & \valpk{0.51} & \valpc{13}$^{\ddagger}$ &  ~ & \valpu{0.04} & \valpk{0.16} & \valpc{300}$^{\ddagger}$ \\
\textsc{~~Recall: Predictions with Some Evidence}   &   ~ & \valpu{0.20} & \valpk{0.24} & \valpc{20}$^{\ddagger}$ &  ~ & \valpu{0.01} & \valpk{0.06} & \valpc{500}$^{\ddagger}$ \\
\midrule
\textsc{~~Law-Paper Evaluations}                  &   ~ & 5739 & 4692 & -- &  ~ & 3199 & 3083 & -- \\
\textsc{~~Papers with Relevant Experiments}       &   ~ & 1493 & 1649 & -- &  ~ & 47 & 221 & -- \\
\textsc{~~Laws with at least 1 relevant paper}    &   ~ & 515 & 467 & -- &  ~ & 34 & 121 & -- \\
\textsc{~~Avg. Papers per Law with Evidence}      &   ~ & 2.9 & 3.5 & -- &  ~ & 1.4 & 1.8 & -- \\
\midrule
\rowcolor[HTML]{E0E0E0}
\multicolumn{9}{l}{\textsc{Self-Assessed Belief (LLM-as-a-judge)}}\\
\midrule
\textsc{~~Likelihood of law being true}             &   ~ & \valpu{0.79} & \valpk{0.83} & \valpc{5}$^\ddagger$ &  ~ & \valpu{0.67} & \valpk{0.76} & \valpc{13}$^\ddagger$ \\
\bottomrule
\end{tabular}
\vspace{-1mm}
\caption{\footnotesize Average \textit{predictive accuracy} of laws as measured by evaluating each law's predictions on recent literature published after the knowledge cutoff used for theory generation. \textit{(top)} Measurements of \textit{predictive precision} and \textit{predictive recall}, evaluated using recent papers. A total of 16,713 evaluations were conducted against 4,554 unique papers (as the same paper can be retrieved when evaluating different but related laws). Definitions of predictive precision and recall are found in Section~\ref{sec:predictive-accuracy}. $\Delta$ represents the relative difference between the parametric-only and literature-supported score, with positive signifying the literature-grounded score is higher. \textit{(bottom)} Average self-assessed likelihood of law plausibility as assessed by the theory generating model. Inferential statistical testing using one-sided non-parametric bootstrap resampling (N=1,000 resamples); significant at $p < 0.01 (\ddagger)$, $p < 0.05 (\dagger)$.} %
\vspace{-2mm}
\label{tab:predictive-accuracy}
\end{table*}

{\flushleft\textbf{Knowledge Window:}} To support the backtesting evaluation paradigm, we use a generation model with a reported knowledge cutoff of June 2024 (\textsc{GPT 4.1}), which provides an 18 month window to the present date at submission time.  The first 12 months of this window are used to supplement the model's own parametric knowledge with  recent literature during theory generation, while papers published in the most recent 6 months are held-out from theory generation, and used for evaluation.

{\flushleft\textbf{Theory Queries from Literature Cross-section:}} To obtain a broad and representative set of theory queries, we randomly sampled 50 papers from recent NLP and AI venues published within several months of the model’s knowledge cutoff, including ACL 2023, EMNLP 2023, AAAI 2023, and NeurIPS 2024.  For each paper, we automatically generated two theory queries, one general and one specific, targeting the paper’s central research theme, yielding a total of 100 theory queries.

{\flushleft\textbf{Theory Generation:}} Using these 100 theory queries, we generated four sets of theories corresponding to each of the two generation conditions: \textit{accuracy-focused} versus \textit{novelty-focused}, and \textit{parametric} versus \textit{literature-supported}.  All subsequent analyses report aggregate properties within these four groups. While the 2,856 total generated theories are not restricted in the number of laws they can contain, each typically contained one or two laws.  To control for variation in theory size, we perform all evaluations at the level of individual laws rather than entire theories.  This resulted in between 790 and 1187 laws per condition. %

\subsection{Hypothesis Generation Metrics}

To assess how access to literature and generation focus affect theory quality, we first use an \textsc{LLM-as-a-Judge} paradigm \cite{NEURIPS2023_91f18a12} to assess theory characteristics relative to the desiderata in Section~\ref{sec:theory-desiderata}, before performing more detailed literature-based analyses for two core metrics, \textit{predictive accuracy} and \textit{novelty}, in the subsequent section. These \textsc{LLM-as-a-judge} metrics are similar to those commonly employed in the ideation literature to assess model-generated research hypotheses \cite{Wang2023SciMONSI,Radensky2024ScideatorHS,vasu-etal-2025-hyper,liu2025hypobenchsystematicprincipledbenchmarking,vasu2025harpatestabilitydrivenliteraturegroundedframework}. %
All ratings are performed on a 10-point Likert scale \cite{likert1932technique}.

Table~\ref{tab:llm-as-a-judge} summarizes results across all conditions.  In the accuracy-focused setting, literature-supported theories outperform parametric theories across nearly all metrics, with particular gains in empirical support and specificity, likely owing to having access to specific details from the literature-derived evidence during theory generation.  In the novelty-focused setting, literature-supported theories are rated as more empirically supported and plausible, though show no gains in specificity, likely owing to the novelty-focused prompt emphasizing making particularly specific predictions to attempt to separate these theories from the high-prior ``safe-zone'' accessed in the accuracy-focused condition.  Across all conditions, model-assessed novelty remains approximately constant.

\subsection{Predictive Accuracy}
\label{sec:predictive-accuracy}
One of the core criteria for evaluating a scientific theory is whether its predictions agree with future empirical results \cite{popper1959logic}.  We operationalize this criterion using a backtesting paradigm: theories are generated using knowledge available up to June 2025, and are then evaluated against empirical results reported in papers published in the subsequent 6 months (July through December 2025).

{\flushleft\textbf{Evaluation Procedure:}}  
Figure~\ref{fig:predictive-accuracy-overview} summarizes our approach for evaluating  predictive accuracy.  At a high level, for each law, a language model first generates a detailed set of empirical predictions that would be expected to hold if the law were true.  We then use \textsc{PaperFinder} to search for recently published papers whose experimental results may speak to these predictions.  Each paper is evaluated independently to determine whether it provides evidence that supports, contradicts, or does not inform each prediction.

For a given prediction, the proportion of supporting versus contradicting evidence serves as a proxy for predictive \textit{precision}.  Similarly, the proportion of laws (or predictions) for which at least one relevant paper can be identified serves as a proxy for predictive \textit{recall}.  Together, these measures capture both how often predictions are empirically supported when tested, and how frequently predictions can be tested using recently published work.

The set of predictions is used as a grading rubric.  Each prediction is represented as a schema that includes (i) the \textit{specific prediction}, and (ii) a set of \textit{operational signals} commonly used to measure that prediction in practice.  For example, if a law predicts that a particular intervention improves machine translation performance, operational signals may include improvements in BLEU \cite{papineni-etal-2002-bleu}, ROUGE \cite{lin-2004-rouge}, or related metrics. To avoid overly permissive matching, each prediction schema also specifies (iii) a \textit{strong test requirement}, describing the concrete experimental comparison a paper must report to be considered a valid test (e.g., reporting machine translation performance both with and without the intervention).  Finally, the schema defines (iv) explicit criteria for what constitutes \textit{supporting} versus \textit{contradicting} evidence.  Taken together, these constraints help impose sharp boundaries on evaluation and work to prevent marginally related results from being misinterpreted as direct evidence.

{\flushleft\textbf{Precision and Recall:}}  
For a given prediction, its precision ($P_{\text{pred}}$) is defined as the number of papers from corpus $C$ providing supporting evidence ($C_{supp}$) divided by the total number of papers from $C$ providing either supporting or contradicting ($C_{supp}$) evidence, i.e., $P_{\text{pred}} = C_{supp} / (C_{supp} + C_{cont})$.  The precision of a law is computed as the average precision of its predictions, excluding predictions for which no evidence is found.  Recall is defined as the proportion of laws for which at least one paper provides direct evidence evaluating one or more of its predictions.  We additionally report recall at the level of individual predictions.

\begin{table*}[t!]
\centering
\footnotesize
\newcolumntype{C}{>{\centering\arraybackslash}p{1.0cm}}
\begin{tabular}{lcCCcCC}
\toprule
~ & ~ & \multicolumn{2}{c}{\textbf{Accuracy-Focused}} & ~ & \multicolumn{2}{c}{\textbf{Novelty-Focused}}    \\
\textbf{Novelty Type}  & ~ & \textbf{Param.} & \textbf{Lit.} & ~ & \textbf{Param.} & \textbf{Lit.} \\
\midrule
\rowcolor[HTML]{E0E0E0}
\multicolumn{7}{l}{\textsc{Qualified Novelty (Relative to papers used for theory generation)}}\\
\midrule
\textsc{~~Phenomenon/Effect}                  &   ~ & \valpu{0.69} & \valpk{0.36} &  ~ & \valpu{0.99} & \valpk{0.90} \\
\textsc{~~Explanation}                        &   ~ & \valpu{0.79} & \valpk{0.49} &  ~ & \valpu{0.96} & \valpk{0.90} \\
\textsc{~~Unification}                        &   ~ & \valpu{0.80} & \valpk{0.68} &  ~ & \valpu{0.97} & \valpk{0.92} \\
\textsc{~~Generalization/Scope Expansion}     &   ~ & \valpu{0.87} & \valpk{0.82} &  ~ & \valpu{0.99} & \valpk{0.96} \\
\textsc{~~Limitation/Scope Constraint}        &   ~ & \valpu{0.42} & \valpk{0.36} &  ~ & \valpu{1.0}  & \valpk{0.96} \\
\textsc{~~Conceptual Reframing/Abstraction}   &   ~ & \valpu{0.76} & \valpk{0.50} &  ~ & \valpu{0.99} & \valpk{0.77} \\
\textsc{~~Meta-Analysis/Empirical Synthesis}  &   ~ & \valpu{0.82} & \valpk{0.70} &  ~ & \valpu{1.0}  & \valpk{0.98} \\
\midrule
\textsc{~~Number of Laws Evaluated}           &   ~ & 100 & 100 & ~ & 100 & 100 \\
\textsc{~~Average Papers Per Evaluation}      &   ~ & 66 & 66 &  ~ & 71 & 71 \\
\textsc{~~Total Papers Evaluated}             &   ~ & 6632 & 6632 &  ~ & 7112 & 7112 \\
\bottomrule
\end{tabular}
\vspace{-2mm}
\caption{\footnotesize \textit{Qualified Novelty Analysis}, with novelty evaluated with respect to the input papers found by \textsc{PaperFinder} when generating a given theory.  Due to pragmatic limitations of cost, each of the four generation conditions is evaluated on 100 randomly selected laws, using a total of 13,744 papers.  Values represent average proportions of laws rated as having a given type of novelty, with higher proportions reflecting more novelty relative to the reference corpus $C$.} %
\vspace{-2mm}
\label{tab:novelty-analysis}
\end{table*}

{\flushleft\textbf{Self-Assessed Belief:}}  
As a complementary signal, we also estimate each law’s plausibility using the generating model’s own self-assessed belief regarding whether its predictions are likely to hold, based solely on parametric knowledge \cite{kadavath2022languagemodelsmostlyknow}.  This self-belief measure provides a model-internal estimate of likelihood, independent of direct literature grounding.  We use the repeated sampling procedure of Agarwal et al.~\cite{agarwal2025autodiscovery} to estimate the probability that a law’s predictions are true.  We report this measure alongside literature-grounded predictive accuracy to contrast the model’s internal beliefs with empirical support observed in the literature.

{\flushleft\textbf{Results:}}  
Predictive accuracy results are shown in Table~\ref{tab:predictive-accuracy}.  \textsc{Claude Sonnet 4.5} was used as the evaluation model.  
On average, 5 predictions were generated per law. Across all conditions, 2,983 laws were evaluated against the full text of 4,554 unique papers published in the most recent 6 months. As the same papers are frequently retrieved when evaluating similar laws, a total of 16,713 law--paper evaluations (5.6 papers per law) were conducted.

In the accuracy-focused condition, laws exhibit similarly high precision across both parametric and literature-supported generation (0.88 and 0.90), with recall ranging from 0.45 to 0.51.  That is, for roughly half of the generated laws, at least one recent paper directly tests one or more of their predictions, and when such tests exist, the evidence overwhelmingly supports the predictions.  This suggests that accuracy-focused theories can achieve broad predictive validity regardless of whether they are generated with explicit literature support, or solely using parametric knowledge.

In contrast, laws generated in the novelty-focused condition exhibit substantially lower predictive accuracy and recall.  Among laws with literature-evaluable predictions, 61\% of predictions from literature-supported theories are supported by recent evidence, compared to only 34\% for parametric theories.  Similarly, recall is substantially lower in this setting: only 16\% of literature-supported laws and 4\% of parametric laws have any directly relevant recent papers.  This pattern is consistent with the novelty-focused prompt producing more uncommon or speculative predictions that are less frequently tested in the current literature.  Taken together, these results indicate that while novelty-focused theories are riskier, access to literature during the theory generation process substantially improves their predictive accuracy when evaluated using literature-grounded backtesting.

Finally, model self-assessed belief aligns well with literature-grounded predictive accuracy in the accuracy-focused condition, but not the novelty-focused condition: the model estimates that 79–83\% of accuracy-focused predictions will hold, which approximately matches the observed empirical precision of 88–90\%.  However, in the novelty-focused condition, the generating model substantially overestimates predictive accuracy (67–76\%) relative to observed support in the literature (34–61\%).  This discrepancy may reflect limitations of backtesting, in that novel predictions may not have been empirically tested in the literature. Similarly, it may also reflect that novelty-focused theories are inherently higher-risk by emphasizing more novel, specific, and higher-entropy laws.

\subsection{Novelty Analysis}
\label{sec:novelty-analysis}

A core desiderata for theories in Section~\ref{sec:theory-desiderata} is the novelty of their contributions relative to existing work.  Exhaustively evaluating novelty against the full scientific literature is impractical at our scale.  We therefore evaluate \textit{qualified novelty}: novelty assessed \textit{with respect to the set of retrieved papers used during theory generation} for a given query.

{\flushleft\textbf{Reference Corpus:}}
Ideally, each law would be evaluated for novelty against the full literature of the field, though this is currently intractible. %
Accordingly, we evaluate novelty relative to the scoped reference corpus $C$ consisting of the $\approx66$--$71$ papers retrieved by \textsc{PaperFinder} for each law’s theory query.  For literature-supported generation, this corpus corresponds directly to the evidence set used during theory synthesis.  For parametric generation, we evaluate novelty against the same set of retrieved papers to enable a matched comparison.  We therefore interpret the resulting scores as \textit{qualified novelty estimates}: they indicate whether a law is novel with respect to the literature most directly relevant to its generation, while acknowledging that overlap with work outside this retrieved set may exist.

{\flushleft\textbf{Novelty Dimensions:}}
We evaluate novelty on seven common dimensions on which scientific ideas are frequently considered as novel. \textsc{Phenomenon/Effect} captures the discovery of a new effect or empirical regularity.  \textsc{Explanation} captures proposing a new mechanism for an existing effect.  \textsc{Unification} captures linking previously separate concepts under a shared framework.  \textsc{Generalization/Scope Expansion} and \textsc{Limitation/Scope Constraint} capture changes to the conditions under which an existing relationship is claimed to hold.  \textsc{Conceptual Reframing/Abstraction} captures reparameterizing or reframing a phenomenon at a different level of description.  Finally, \textsc{Meta-analysis/Empirical Synthesis} captures new insights obtained by synthesizing results across prior empirical studies.

{\flushleft\textbf{Evaluation Procedure:}}
Evaluation proceeded similarly to predictive accuracy.  For each law, we evaluate novelty relative to each retrieved reference paper in $C$ using a per-paper judge model.  The resulting per-paper novelty assessments are then provided to a consolidating judge model, which produces a final decision for each novelty dimension.  We evaluate each dimension independently (i.e. 7 consolidation calls per law) and sort per-paper results in order of descending novelty to conservatively present evidence against novelty in the portion of the context window where the model is most sensitive \cite{liu-etal-2024-lost}.  \textsc{GPT-5-mini} is used as the per-paper judge model and \textsc{Claude Sonnet 4.5} as the consolidating judge model. Due to evaluation cost ($\approx\$3$ per law, per condition), we evaluate a randomly selected subset of 100 laws per condition.

\begin{figure}[t!]
    \centering
    \includegraphics[scale=0.47]{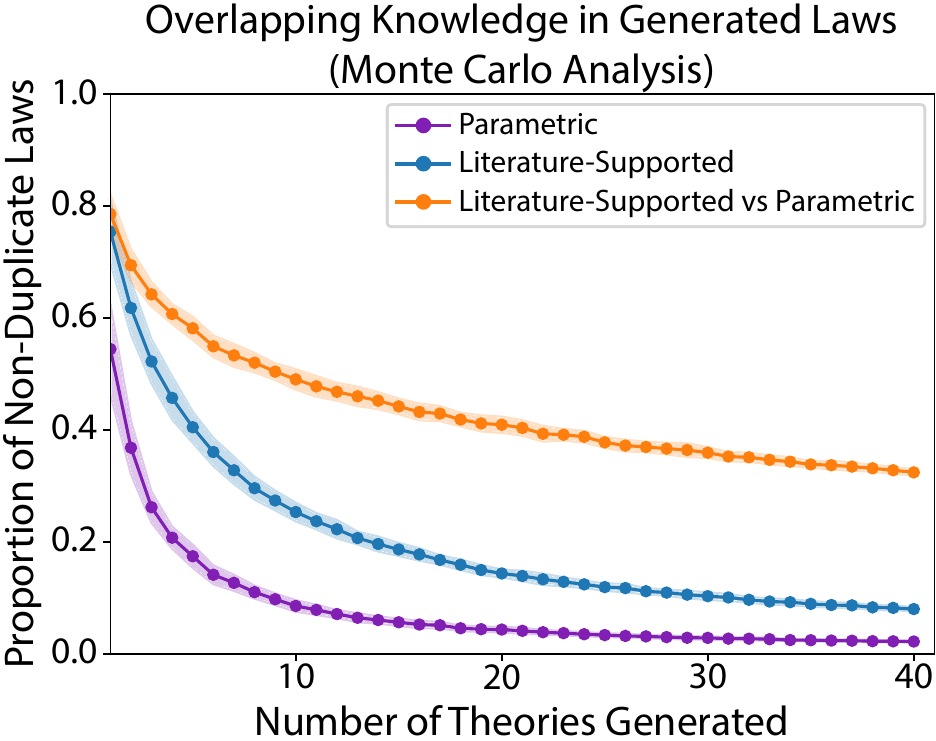}
    \caption{\footnotesize Monte Carlo analysis of theory law overlap when repeatedly generating theories using the same theory query.  Parametric and literature-supported series measure duplicates \textit{within} group (i.e. randomly select a \textit{parametric} theory, then check whether it is duplicated in a random sample of N \textit{parametric} theories). The literature-supported vs parametric series measures duplicates \textit{across} groups (i.e. randomly select a \textit{parametric} theory, then check wither it is duplicated in a random sample of N \textit{literature-supported} theories), and is designed to measure whether literature-supported vs parametric theories are generating similar content. Duplication is measured using a pairwise \textsc{LLM-as-a-judge} using two input laws. Each point reflects the average of 50 samples. Shaded areas represent standard deviation across the 50 samples. Additional details in the \textsc{Appendix}.}    
    \label{fig:knowledge-overlap-analysis}
\end{figure}

{\flushleft\textbf{Results:}}
Qualified novelty results are shown in Table~\ref{tab:novelty-analysis}. Across conditions, the system frequently generates laws that are judged novel relative to the reference papers used during theory generation, suggesting a nontrivial capacity to synthesize ideas not explicitly present in the retrieved literature.  In the accuracy-focused setting, novelty most often arises through \textsc{Generalization/Scope Expansion} and least often through discovery of new \textsc{Phenomena/Effects}.  Literature-supported laws are generally rated as less novel than parametric laws, consistent with anchoring effects from explicit access to retrieved evidence during generation, or alternatively, a genuine capacity for pretrained models to synthesize novel contributions from parametric knowledge alone.  In the novelty-focused setting, laws are rated as highly novel across nearly all dimensions, reflecting the prompt’s emphasis on novel, specific, high-entropy theory generation.

\section{Discussion}

{\flushleft\textbf{Novelty versus Predictive Accuracy.}}
Across our experiments, novelty and predictive accuracy exhibit a clear tradeoff.  The qualified novelty analysis shows that the \textit{novelty-focused} condition produces highly novel laws relative to the reference literature.  However, novelty in this setting reflects absence from the retrieved papers, not empirical validity.  Consistent with this distinction, the predictive accuracy results indicate that novelty-focused laws are substantially riskier when evaluated using literature-based backtesting.  While a subset of these laws may represent genuinely new and correct hypotheses, others are likely novel because they are speculative or inconsistent with empirical results reported in recent work.  Taken together, these findings suggest that novelty-focused generation expands the space of candidate theories, but at the cost of reduced reliability under a literature-grounded backtesting-style evaluation.

{\flushleft\textbf{Parametric Saturation and Literature-Guided Exploration.}}
A desirable property of a theory generation system is for it to be able to generate a diverse set of theories for a given theory query.  To investigate how different knowledge sources affect the diversity of generated theories, we conducted a monte-carlo analysis that shows the proportion of duplicate theories that are generated when the same theory query is repeatedly submitted. This analysis, shown in Figure~\ref{fig:knowledge-overlap-analysis}, shows that theories generated with parametric knowledge alone quickly saturate to generating only duplicates, mirroring observations in hypothesis generation \cite{si2025can}, while also showing theories generated with knowledge from literature have a lower proportion of duplicates.  Critically, when comparing theories generated \textit{across} conditions rather than \textit{within} conditions --- that is, comparing literature-supported theories against theories generated with parametric knowledge --- the proportion of non-duplicate statements after generating 40 theories is 32\%, suggesting that literature-supported and parametric theories are generating laws with different content in a large proportion of cases.

\section{Conclusion}
We formulate the task of synthesizing theories from corpora of research papers, demonstrate a system that generates thousands of candidate theories from tens of thousands of input papers, and analyze how theory characteristics vary under accuracy-focused or novelty-focused objectives, with either literature support or parametric knowledge alone. Our literature-supported generation method, though more expensive, creates theories with significantly greater specificity, greater empirical support, and higher predictive accuracy when compared with subsequently reported results.  In future work we hope to develop further improved techniques for theory synthesis, including methods that yield higher {\em novelty}.%

\section{Limitations}

{\flushleft\textbf{Backtesting:}} We evaluate predictive accuracy using the backtesting paradigm \cite{YETISGENYILDIZ2009633}.  Nominally, when a theory makes experimental predictions, these are typically evaluated by conducting a series of new experiments to find evidence that supports or contradicts those predictions.  Experimental science is expensive in both time and cost, and single experiments can take months, and cost thousands of dollars and other resources.  Backtesting avoids much of this expense by holding out the most recent literature during training, and using this as an evaluation set.  The central trade-off is in recall: the number of papers that happen to have been published that test the specific predictions a given theory makes will be small.  In this work we observe that approximately 51\% of \textit{accuracy-focused} theories have at least one paper that tests its laws predictions, though this reduces substantially to only 4\% of theories in the \textit{novelty-focused} condition. As a result, many of the predictions that the model-generated theories make are not fully tested in the literature. Further, literature has a positive-result bias: negative results often go unreported, which may make finding contradictory evidence more challenging through literature search.  An individual researcher using model-generated theories to help generate novel experimental predictions to test may be able to reasonably test an individual theory with a normative number of experiments, but theory generation at scale pragmatically requires more scalable methods of evaluation, such as backtesting.  The automation of some kinds of experimental science may allow some theories to be experimentally evaluated partially automatically in the near future.  However, current experimental model performance can be low -- for example, \textsc{CodeScientist} \cite{jansen-etal-2025-codescientist} reports that when its discoveries were examined by a human domain expert, only 30\% of automatic discoveries passed a code review and replication.

{\flushleft\textbf{Cost:}} Theory generation and (particularly) evaluation is currently somewhat expensive, limiting the capacity to scale.  An estimate of \textsc{LLM}-associated \textsc{API} costs is provided in the \textsc{Appendix} (See Table~\ref{tab:cost-analysis}).  Similarly, the theory generation pipeline takes approximately 15-30 minutes per theory query, though this can be parallelized up to \textsc{API} rate limits for LLM calls and paper retrieval limits.

{\flushleft\textbf{Open Access Literature:}} Currently our theory generation scope is limited to fields whose scholarly literature is primarily open access and available for automated download, such as artificial intelligence and natural language processing.  Similarly, searching for and downloading these source papers for generation and evaluation is the primary time component of our system, largely due to rate limits in searching for and downloading literature.

\bibliography{anthology_part1,anthology_part2,custom}

\appendix

\section{Supplementary Material}
The supplementary material includes: 

{\flushleft\textbf{Theories:}} An example set of theories, extraction schema, extraction results, predictive accuracy evaluations, and qualified novelty evaluations.

{\flushleft\textbf{Code:}} The full release will include the \textsc{Theorizer} code (on \textsc{Github}), which includes a user interface, \textsc{API}, and all prompts. 

\section{Follow-Through Example}

While the supplementary material includes hundreds of model-generated theories, schemas, and evaluations in easily-viewable \textsc{HTML} format, the following follow-through example on Intelligent Tutoring Systems is provided in the \textsc{Appendix}: 

\begin{enumerate}[noitemsep, topsep=0pt]
    \item Generated Theory: Table~\ref{tab:example1-law}
    \item Predictive Accuracy Evaluation: Table~\ref{tab:example1-law-predictive-accuracy}
    \item Qualified Novelty Evaluation: Tables~\ref{tab:example1-law-novelty} and \ref{tab:example1-law-novelty1}
    \item Extraction Schema: Table~\ref{tab:extraction-schema-example}
    \item Extraction from Paper: Table~\ref{tab:extraction-example}
\end{enumerate}

\section{Theory Generation Hyperparameters}
The theory generation procedure is described in Section~\ref{sec:theory-generation}, with full details including promtps in the code release. 

{\flushleft\textbf{Models:}} The generation models are user-selectable in the \textsc{Theorizer} user interface and \textsc{API}. In the experiments reported here, the generation model was \textsc{GPT-4.1}, which is used for all aspects of theory generation (i.e. extraction schema generation, theory generation, theory reflection) except for extracting evidence from individual papers, which is performed by an inexpensive model (\textsc{GPT-5-mini}) due to the scale required.  Generation temperature was set to \textsc{0} in the high-accuracy condition, and \textsc{1} in the novelty-focused condition. 

{\flushleft\textbf{Self-Evaluated Novelty and Filtering:}} Independent of the qualified novelty evaluation in Section~\ref{sec:novelty-analysis}, the theory generation process itself requires generated theories to include self-assessments of novelty provided by the generating model, to encourage the model to generate laws that are not repetitions of existing/well-known laws. This assessment includes a chain-of-thought style assessment of the novelty (\textit{what already exists, what is novel, provide an explanation/reasoning as to how the novelty of this law could be classified}), followed by a categorical assessment of the law's novelty using one of 4 classification labels.  Those labels, in order of decreasing novelty, are: \textsc{1. new, 2. somewhat-related-to-existing, 3. closely-related-to-existing, 4. existing}.  To focus our evaluation primarily on novel theories/laws, we filtered any laws rated as \textsc{closely-related-to-existing} or \textsc{existing} from further analysis.  The proportion of these laws filtered varied between 1\% and 18\%, depending on generation condition.  The total number of laws reported in Table~\ref{tab:llm-as-a-judge} is the total after this filtering.

{\flushleft\textbf{Subsampling Evidence:}} Depending on the number of papers retrieved and the amount of evidence in each paper, the total size (in tokens) of all retrieved evidence can extend beyond the generating model's input context token limit. When this happens, the evidence is randomly subsampled. 

{\flushleft\textbf{Evaluation Models:}} Unless otherwise stated, the evaluation model for all evaluations was \textsc{Claude Sonnet 4.5}.  In cases where a large number of individual evaluations was required (e.g. evaluations against a corpus of papers), this model is sometimes augmented with a more inexpensive model, \textsc{GPT-5-mini}, and these cases are described in text.

\section{Generation Costs}

Estimated \textsc{LLM API} costs are included in Table~\ref{tab:cost-analysis}.

\begin{table}[h!]
\centering
\footnotesize
\begin{tabular}{ll}
\toprule
\textbf{Activity} &  \textbf{Cost Estimate}      \\
\midrule
\rowcolor[HTML]{E8E8E8}
Theory Generation       &  ~  \\
~~~\textit{Literature-Supported}  &  $\$0.26$ per theory   \\ 
\rowcolor[HTML]{E8E8E8}
~~~\textit{Parametric Only}       &  $\$0.04$ per theory   \\ 
LLM-as-a-Judge Metrics  &  $\$0.03$ per law \\
\rowcolor[HTML]{E8E8E8}
Predictive Accuracy     &  $\$0.39$ per law ($\approx5$ papers) \\
Surprisal Analysis     &  $\$0.03$ per law \\
\rowcolor[HTML]{E8E8E8}
Novelty Evaluation      &  $\$2.46$ per law ($\approx70$ papers) \\
Overlap Analysis        &  $\$10$ per law \\
\bottomrule
\end{tabular}
\caption{\footnotesize Estimated LLM \textsc{API} costs for the generation and evaluation of theories in this work.}
\label{tab:cost-analysis}
\end{table}

\begin{table*}[t]
\centering
\scriptsize
\setlength{\tabcolsep}{6pt}
\renewcommand{\arraystretch}{1.15}
\begin{tabular}{p{0.15\textwidth} p{0.80\textwidth}}
\toprule

\multicolumn{2}{c}{\textbf{Example Theory: Emergent Multi-Agent Social Presence Theory in LLM-ITS (Intelligent Tutoring Systems)}} \\
\midrule
\rowcolor[HTML]{E8E8E8}
\multicolumn{2}{l}{\textbf{Theory Query}} \\
\midrule

\textbf{Theory Query} & Build a theory of how the explicit integration of step-by-step scaffolding and dynamic conversational strategies, as operationalized in the CLASS framework, influences student cognitive gains, engagement, and motivation in LLM-powered intelligent tutoring systems across diverse subject domains.\\

\midrule
\rowcolor[HTML]{E8E8E8}
\multicolumn{2}{l}{\textbf{Theory (General)}} \\
\midrule

\textbf{Generation Condition} & \textit{literature-supported, accuracy-focused} \\
\textbf{Theory ID} & theory-215 \\
\textbf{Theory Name} & Emergent Multi-Agent Social Presence Theory in LLM-ITS (Intelligent Tutoring Systems)\\
\textbf{Theory Description} & This theory asserts that the explicit integration of multi-agent conversational strategies (e.g., simulated classmates, assistants, note-takers) in LLM-powered ITS can create emergent social presence and collaborative learning dynamics, which in turn enhance engagement, motivation, and, under certain conditions, cognitive gains. The effect is mediated by the diversity of agent roles, the orchestration of turn-taking, and the degree of user participation. However, the benefits are domain- and context-dependent, and excessive agent complexity or poorly managed interaction can reduce learning efficiency. \\

\midrule
\rowcolor[HTML]{E8E8E8}
\multicolumn{2}{l}{\textbf{One Law/Statement}} \\
\midrule
\textbf{Law Name} & Multi-Agent Social Presence Law \\
\textbf{Law Statement} & In LLM-powered ITS, the presence of multiple simulated agent roles (e.g., teacher, assistant, classmates) and dynamic conversational orchestration increases perceived social presence, engagement, and motivation, as measured by interaction metrics and self-report, compared to single-agent or non-interactive systems. \\
\textbf{Law Type} & \textit{qualitative} \\
\textbf{Scope/Domain} & LLM-powered ITS and classroom simulations with multi-agent conversational architectures, across higher education and K-12 domains, especially in settings where social/collaborative learning is valued. \\
\textbf{Special Cases} & 1. In highly individual or assessment-driven tasks, social presence may have less impact on cognitive gains.\\
 & 2. For learners with social anxiety, increased agent interaction may not increase motivation. \\

\midrule

\textbf{Supporting Evidence} &
1. SimClass multi-agent classroom ablation showed that removing classmate agents reduced user speech length by 26.5\% (TAGI) and 45.2\% (HSU), and reduced Community of Inquiry (CoI) social and cognitive presence scores; full multi-agent systems had higher engagement and learning experience. (UUIDs: e2892.0)\\
& 2. SimClass FIAS coding showed high Student Initiation Ratios (SIR ~0.9), indicating active participation in multi-agent settings. (UUIDs: e2892.0)\\
& 3. SRLAgent's gamified, multi-agent orchestration (Planning Agent, SubTask Tutor, Reflection Agent) increased engagement and SRL skills compared to baseline multimedia learning. (UUIDs: e2717.0) \\
& 4. EnglishBot's open conversational practice (simulated dialogue) led to greater engagement and learning gains than a listen-and-repeat interface. (UUIDs: e2760.3) \\
& 5. DBTS (Discussion-Based Teaching Systems) report ~72\% increase in engagement and ~74\% improvement in learning outcomes, attributed to dialogic, multi-agent interaction. (UUIDs: e2771.2) \\

\midrule
\rowcolor[HTML]{E8E8E8}
\multicolumn{2}{l}{\textbf{Self-Assessed Law Novelty} \textit{(produced as part of theory generation; independent of later novelty evaluation)}} \\
\midrule
\textbf{What Already Exists} & Social presence and collaborative learning are established in educational theory, and multi-agent systems have been explored in ITS, but the emergent effects of LLM-powered multi-agent orchestration on engagement and learning are not formalized. \\
\textbf{What Is Novel} & The explicit law linking multi-agent LLM orchestration to emergent social presence and its downstream effects in ITS is novel. \\
\textbf{Classification Expl.} & Existing work discusses social presence and collaborative learning, but not the emergent effects of LLM-driven multi-agent orchestration in ITS. \\
\textbf{Classification} & \textit{new} \\
\textbf{LLM-Generated Refs.} &
1. Garrison et al. (2000) Critical Inquiry in a Text-Based Environment: Computer Conferencing in Higher Education [Community of Inquiry framework] \\
& 2. Dillenbourg (1999) Collaborative learning: Cognitive and computational approaches [collaborative learning, not LLM multi-agent orchestration] \\

\bottomrule
\end{tabular}
\caption{\footnotesize An example theory generated in this work, including the theory name, description, a single law, and a self-assessment of novelty made from the generation model.  An example predictive accuracy evaluation for this theory is provided in Table~\ref{tab:example1-law-predictive-accuracy}, and the qualified novelty evaluation relative to the source papers used to generate this theory is in Table~\ref{tab:example1-law-novelty}.  The extraction schema used to generate this theory is in Table~\ref{tab:extraction-schema-example}, and example extracted evidence shown in Table~\ref{tab:extraction-example}. Hundreds of additional examples are available in \textsc{HTML} format in the Supplementary Material.}
\label{tab:example1-law}
\end{table*}

\begin{table*}[t]
\centering
\scriptsize
\setlength{\tabcolsep}{6pt}
\renewcommand{\arraystretch}{1.15}
\begin{tabular}{p{0.15\textwidth} p{0.80\textwidth}}
\toprule

\multicolumn{2}{c}{\textbf{Predictive Accuracy Example: Emergent Multi-Agent Social Presence Theory in LLM-ITS (Intelligent Tutoring Systems)}} \\
\midrule
\rowcolor[HTML]{E8E8E8}
\multicolumn{2}{l}{\textbf{Evaluation (General)}} \\
\midrule

\textbf{Theory ID} & theory-215 \\
\textbf{Theory Name} & Emergent Multi-Agent Social Presence Theory in LLM-ITS \\
\textbf{Theory Statement} & In LLM-powered ITS, the presence of multiple simulated agent roles (e.g., teacher, assistant, classmates) and dynamic conversational orchestration increases perceived social presence, engagement, and motivation, as measured by interaction metrics and self-report, compared to single-agent or non-interactive systems. \\
\textbf{Papers Examined} & 14 papers published between July 2025-December 2025 were examined.\\
\textbf{Average Precision} & 1.0 \textit{(of the 2 rubric items with some evidence found)}\\

\midrule
\rowcolor[HTML]{E8E8E8}
\multicolumn{2}{l}{\textbf{Rubric Evaluation (Per Prediction) \textit{(Note: Only 3 of 5 rubric items shown for space)}}} \\
\midrule

\textbf{Prediction Short Name} & agent\_role\_diversity\_effect \\
\textbf{Specific Prediction} & Within multi-agent LLM-ITS, increasing the diversity of agent roles (e.g., adding classmate agents to teacher+assistant) increases engagement or social presence compared to reduced-role configurations. \\
\textbf{Operational Signals} & 1. Ablation studies varying number or types of agent roles \\
 & 2. Engagement or social presence metrics across different role configurations \\
 & 3. Explicit manipulation of agent diversity (e.g., teacher-only vs teacher+classmates) \\
\textbf{Strong Test Requirement} & A paper must report an ablation or comparison where agent role diversity is systematically varied, with engagement or social presence outcomes measured for each configuration. \\
\textbf{What Does Support Look Like} & Adding diverse agent roles (e.g., classmates) significantly increases engagement metrics or social presence scores compared to fewer-role configurations. \\
\textbf{What Does Contradiction Look Like} & Removing or adding agent roles has no significant effect on engagement or social presence, or reduced-role systems perform equivalently or better. \\
\textbf{Counts} & support: 1 paper, contradict: 0 papers, no\_evidence: 13 papers\\
\textbf{Proportions} & support: 1.0, contradict: 0.0 \\
\textbf{Evidence} & 1. \textit{Source Paper:} OnlineMate: An LLM-Based Multi-Agent Companion System for Cognitive Support in Online Learning (Xian Gao, Zongyun Zhang, Ting Liu, Yuzhuo Fu; 2025/9; ArXiv: 2509.14803) \\
& \textit{Evidence Quote or Locator:} Figure 5 and associated text: 'Figure 5 illustrates the impact of varying the number of OnlineMate Agents on the highest cognitive level attained by students. As the teaching assistant role transitions from a single agent to multiple agents engaged in classroom discussions, a noticeable improvement in students' highest cognitive levels is observed... However, when the number of agents exceeds four, the increase in cognitive level begins to plateau, and a decline is even noted when the number of agents surpasses six.' \\
& \textit{Evidence in Support:} The paper reports an ablation study systematically varying the number of agents (from 1 to 6+) and measuring cognitive level outcomes. Results show that increasing from single to multiple agents (up to 4) produces noticeable improvements in cognitive levels, directly supporting the prediction that agent role diversity increases engagement/outcomes. \\

\midrule

\textbf{Prediction Short Name} & assessment\_task\_boundary \\
\textbf{Specific Prediction} & In highly individual or assessment-driven tasks (e.g., standardized testing, individual skill drills), multi-agent social presence has reduced or no significant impact on cognitive learning gains (e.g., test scores, knowledge assessments) compared to collaborative or exploratory learning contexts. \\
\textbf{Operational Signals} & 1. Comparison of multi-agent effects across task types: assessment-driven vs collaborative/exploratory \\
 & 2. Learning outcome measures (test scores, knowledge gains) in assessment contexts \\
 & 3. Explicit categorization of tasks as individual/assessment vs collaborative \\
\textbf{Strong Test Requirement} & A paper must compare multi-agent vs single-agent LLM-ITS effects on cognitive outcomes in an explicitly assessment-driven or individual task context, with clear outcome measures. \\
\textbf{What Does Support Look Like} & Multi-agent presence shows minimal or no significant advantage on cognitive gains in assessment-driven tasks, even if engagement increases. \\
\textbf{What Does Contradiction Look Like} & Multi-agent systems produce significant cognitive gains in highly individual or assessment-driven tasks comparable to those in collaborative contexts. \\
\textbf{Counts} & support: 0 papers, contradict: 0 papers, no\_evidence: 14 papers \\
\textbf{Proportions} & support: 0, contradict: 0 \\
\textbf{Evidence} & \textit{No evidence found in recent papers (last 6 months) that test this prediction.}\\

\midrule

\textbf{Prediction Short Name} & excessive\_complexity\_inefficiency \\
\textbf{Specific Prediction} & LLM-powered ITS with excessive agent complexity or poorly managed interaction (e.g., unclear turn-taking, overwhelming number of agents, confusing orchestration) reduces learning efficiency, as measured by time-on-task, task completion rates, or learning gains per unit time, compared to well-orchestrated multi-agent systems. \\
\textbf{Operational Signals} & 1. Comparison of different orchestration strategies or agent complexity levels \\
 & 2. Learning efficiency metrics: time to mastery, learning gains per minute, task completion rates \\
 & 3. User confusion, cognitive load, or negative feedback related to agent interaction complexity \\
\textbf{Strong Test Requirement} & A paper must compare different levels of agent complexity or orchestration quality in multi-agent LLM-ITS, measuring learning efficiency or related outcomes. \\
\textbf{What Does Support Look Like} & Overly complex or poorly orchestrated multi-agent systems show reduced efficiency metrics (longer time, lower completion rates, or reduced gains per unit time) compared to simpler or well-managed configurations. \\
\textbf{What Does Contradiction Look Like} & Increased agent complexity consistently improves or has no negative effect on learning efficiency, regardless of orchestration quality. \\
\textbf{Counts} & support: 1 paper, contradict: 0 papers, no\_evidence: 13 papers \\
\textbf{Proportions} & support: 1.0, contradict: 0.0 \\
\textbf{Evidence} & 1. \textit{Source Paper:} OnlineMate: An LLM-Based Multi-Agent Companion System for Cognitive Support in Online Learning (Xian Gao, Zongyun Zhang, Ting Liu, Yuzhuo Fu; 2025/9; ArXiv: 2509.14803) \\
& \textit{Evidence Quote or Locator:} Figure 5 and associated text: 'However, when the number of agents exceeds four, the increase in cognitive level begins to plateau, and a decline is even noted when the number of agents surpasses six. This suggests that, in online learning contexts, an excessive number of agent companions may not necessarily be beneficial. An overabundance of agents providing excessive information may lead to cognitive overload, hindering the students'' \\
& \textit{Evidence in Support:} The paper directly tests varying levels of agent complexity (number of agents) and finds that excessive agents (>6) lead to declining cognitive outcomes, explicitly attributing this to cognitive overload from too much information. This directly supports the prediction that excessive complexity reduces learning efficiency. \\

\bottomrule
\end{tabular}
\caption{\footnotesize Example Predictive Accuracy evaluation, including the rubric, and supporting evidence for two rubric elements.}
\label{tab:example1-law-predictive-accuracy}
\end{table*}

\begin{table*}[t]
\centering
\scriptsize
\setlength{\tabcolsep}{6pt}
\renewcommand{\arraystretch}{1.15}
\begin{tabular}{p{0.125\textwidth} p{0.80\textwidth}}
\toprule

\multicolumn{2}{c}{\textbf{Qualified Novelty Evaluation: Emergent Multi-Agent Social Presence Theory in LLM-ITS (Intelligent Tutoring Systems)}} \\
\midrule
\rowcolor[HTML]{E8E8E8}
\multicolumn{2}{l}{\textbf{Evaluation (General)}} \\
\midrule
\textbf{Theory ID} & theory-215 \\
\textbf{Theory Name} & Emergent Multi-Agent Social Presence Theory in LLM-ITS \\
\textbf{Theory Statement} & In LLM-powered ITS, the presence of multiple simulated agent roles (e.g., teacher, assistant, classmates) and dynamic conversational orchestration increases perceived social presence, engagement, and motivation, as measured by interaction metrics and self-report, compared to single-agent or non-interactive systems. \\
\textbf{Papers Evaluated} & 54 papers \textit{(that were used as literature-support in the generation of this theory)} \\
\midrule
\rowcolor[HTML]{E8E8E8}
\multicolumn{2}{l}{\textbf{Final Aggregated Qualified Novelty Evaluation (Per Dimension)}} \\
\midrule

\textbf{Dimension} & phenomenon\_effect \\
\textbf{What Is Known} & Three to four recent papers (Ruffle\&Riley studies and SimClass, 2023-2024) provide direct experimental evidence that multi-agent LLM conversational architectures with distinct simulated roles (teacher, assistant, classmates) produce statistically significant increases in self-reported engagement, perceived social presence (CoI framework), and interaction metrics compared to single-agent or non-interactive baselines. These studies report quantitative effect sizes across multiple validated instruments (CoI subscales, FIAS coding, message counts, satisfaction scores) in controlled experiments and semester-long deployments, establishing the phenomenon empirically within their specific implementations. \\
\textbf{What Introduced} & The Multi-Agent Social Presence Law articulates this as a generalizable empirical regularity: that explicit integration of multiple simulated agent roles with dynamic conversational orchestration systematically produces emergent social presence and collaborative learning dynamics that increase perceived engagement and motivation across LLM-powered ITS contexts. The theory names and frames this as a distinct, measurable phenomenon mediated by role diversity, turn-taking orchestration, and user participation, extending beyond isolated implementations to propose a general principle. \\
\textbf{What Novel} & While the core phenomenon has been explicitly demonstrated with rigorous experimental evidence in 3-4 papers, the overwhelming majority of the LLM-ITS literature (43/59 papers assessed) does not study, measure, implement, or discuss multi-agent social presence as a distinct effect. Most work focuses on single-agent personalization, hint generation, or pedagogical strategies without investigating role-diverse architectures or emergent social-presence dynamics. The phenomenon exists as an explicitly reported finding in isolated recent studies but has not achieved field-wide recognition, systematic replication across contexts, or treatment as established knowledge; experts outside these specific research groups would likely not cite it as a known regularity. \\
\textbf{Degree of Novelty} & explicit\_peripheral \\

\midrule

\textbf{Dimension} & explanatory\_mechanistic \\
\textbf{What Is Known} & The literature documents various mechanisms in isolation or within single-agent paradigms: LLM tutors employ prompt engineering, scaffolding, and adaptive feedback; traditional ITS use learner modeling and personalization; some systems implement multi-agent architectures with role-specific prompts and orchestration (2 papers peripherally mention related components); social presence and engagement are recognized constructs. However, these elements are analyzed separately, and no paper articulates a unified causal mediation model linking multi-agent architectural features to social-psychological mediators to learning outcomes. \\
\textbf{What Introduced} & The theory proposes a novel causal mediation framework specific to multi-agent LLM-powered ITS: diversity of agent roles (teacher, peers, assistants), orchestration of conversational turn-taking, and degree of user participation act as joint architectural mediators that produce emergent social presence as a psychological mechanism, which in turn causally increases engagement, motivation, and (under boundary conditions) cognitive gains. This articulates an integrated causal chain: multi-agent architecture $\rightarrow$ emergent social presence $\rightarrow$ affective/motivational outcomes $\rightarrow$ conditional cognitive effects. \\
\textbf{What Novel} & Across 54 papers, the vast majority (44 papers, 81\%) propose entirely different mechanisms (single-agent scaffolding, prompt engineering, personalization algorithms) or do not address multi-agent social dynamics at all. Two papers mention related mechanistic components peripherally but do not formalize them as causal models; eight papers acknowledge the components are derivable but unstated. Critically, no paper in the corpus articulates or tests the specific causal mediation structure proposed by the theory---that multi-agent role diversity and orchestration jointly produce emergent social presence as the proximate cause of engagement and learning gains. This represents a genuinely new explanatory framework that synthesizes previously disconnected architectural and psychological constructs into a testable causal theory. \\
\textbf{Degree of Novelty} & genuinely\_new \\

\midrule

\textbf{Dimension} & unification \\
\textbf{What Is Known} & The literature contains separate strands discussing ITS architectures, virtual agents, social presence theory in online learning, and collaborative learning dynamics. Multiple systematic reviews and surveys synthesize these areas and provide taxonomies (e.g., three-class tutoring-system taxonomy, pedagogical frameworks, LLM-education taxonomies), and several papers unify technical components within single-agent systems (e.g., LLM+ITS integration, student modeling, prompt engineering). However, no reviewed paper explicitly elevates multi-agent conversational architectures as a unifying design principle that subsumes engagement and motivation outcomes under an emergent social-presence framework. \\
\textbf{What Introduced} & The theory explicitly unifies multi-agent conversational strategies (simulated teacher, assistant, classmates, note-takers) and their orchestration as instances of a single underlying principle: that role diversity and turn-taking dynamics in LLM-powered ITS produce emergent social presence, which causally drives measurable improvements in engagement, motivation, and collaborative learning dynamics across educational contexts. This unification treats previously separate phenomena (agent-based tutoring, social presence effects, collaborative learning gains) as manifestations of one coherent mechanism. \\
\textbf{What Novel} & While component ideas exist separately across the literature---multi-agent systems in education, social presence theory, ITS engagement measures, collaborative learning benefits---the explicit unifying principle that diverse multi-agent conversational role architectures are manifestations of a single emergent social-presence mechanism is not stated in any reviewed paper. This specific conceptual unification could be derived by synthesizing insights across multiple papers (especially reviews discussing agents, ITS design, engagement theory, and collaborative learning), but it has not been explicitly formulated, validated, or presented as a unified theoretical principle in prior work. The shift from separate strands to an integrated explanatory framework represents a novel synthesis. \\
\textbf{Degree of Novelty} & derivable\_unstated \\

\midrule
\rowcolor[HTML]{E8E8E8}
\multicolumn{2}{l}{\textbf{Continued in Table~\ref{tab:example1-law-novelty1}}} \\
\bottomrule
\end{tabular}
\caption{\footnotesize Example Qualified Novelty aggregated evaluation on 7 novelty dimensions.  Due to space limitations, individual evaluations of the 54 source papers examined to create this evaluation are not shown, and evaluation for the remaining 4 novelty dimensions continues on Table~\ref{tab:example1-law-novelty1}.}
\label{tab:example1-law-novelty}
\end{table*}

\begin{table*}[t]
\centering
\scriptsize
\setlength{\tabcolsep}{6pt}
\renewcommand{\arraystretch}{1.15}
\begin{tabular}{p{0.115\textwidth} p{0.83\textwidth}}
\toprule

\multicolumn{2}{c}{\textbf{Qualified Novelty Evaluation (Continued): Emergent Multi-Agent Social Presence Theory in LLM-ITS (Intelligent Tutoring Systems)}} \\
\midrule

\textbf{Dimension} & generalization\_scope\_expansion \\
\textbf{What Is Known} & The surveyed literature (54 papers) demonstrates LLM-based educational effectiveness in narrow, specific contexts: predominantly single-agent architectures evaluated within particular domains (programming, mathematics, language learning), specific educational levels (individual courses or grade levels), and limited populations. Multiple review papers discuss broad AI/ITS applicability across educational settings but do not establish that multi-agent conversational architectures produce consistent effects. Only one paper (Simulating Classroom Education) implements multi-agent LLM classroom simulation, but strictly in two university courses with explicit limitations on generalizability; 53 other papers use single-agent designs or do not address multi-agent architectures at all. \\
\textbf{What Introduced} & The theory simultaneously generalizes across three independent axes: (1) architectural: from single-agent tutors to multi-agent conversational systems with diverse simulated roles (teacher, assistant, classmates, note-takers) and dynamic turn-taking orchestration; (2) educational scope: across K-12 and higher education levels; (3) domain scope: across contexts where social/collaborative learning is valued. It asserts these multi-agent architectures produce emergent social presence, engagement, and motivation benefits compared to single-agent or non-interactive systems, with specified boundary conditions (e.g., assessment-driven tasks may show reduced benefits, learners with social anxiety may not benefit from increased interaction). \\
\textbf{What Novel} & The aggregated evidence shows 31/54 papers (57\%) rate this scope expansion as genuinely new relative to their work, 21/54 (39\%) as derivable but unstated, and only 2/54 (4\%) as explicit but peripheral. The vast majority explicitly limit claims to their specific domains, populations, or single-agent architectures and provide no empirical or analytical evidence for the theory's three-fold generalization. The claim that multi-agent conversational architectures in LLM-powered ITS produce consistent cross-domain social-presence and engagement effects across educational levels is not established, demonstrated, or derived in the surveyed literature; while plausible extensions of component findings exist, their unification into a broad multi-agent, cross-context law represents a genuine scope expansion. \\
\textbf{Degree of Novelty} & genuinely\_new \\

\midrule

\textbf{Dimension} & constraint\_limitation \\
\textbf{What Is Known} & Three Ruffle\&Riley papers explicitly establish that multi-agent conversational ITS can increase engagement and perceived social presence without corresponding short-term learning gains, while incurring time costs and interaction quality problems (lenient feedback, repeated prompts). Broader literature documents general LLM/ITS limitations (hallucinations, bias, over-reliance risks, need for human oversight) but primarily for single-agent systems. Context-dependence of educational technology effectiveness across learner populations and task types is widely acknowledged as a general principle. \\
\textbf{What Introduced} & The theory explicitly enumerates a systematic set of multi-agent-specific constraints: (1) benefits are domain- and context-dependent, with particular reductions in highly individual or assessment-driven tasks; (2) excessive agent complexity or poorly managed multi-agent orchestration constitutes a distinct failure mode that reduces learning efficiency; (3) learner traits, specifically social anxiety, moderate or reverse motivational benefits of increased agent interaction. These are framed as necessary boundary conditions for multi-agent conversational architectures rather than general LLM caveats. \\
\textbf{What Novel} & The basic dissociation between engagement and learning gains is empirically established by Ruffle\&Riley work. However, the identification of *excessive multi-agent complexity* and *orchestration failures* as specific, nameable failure modes is not explicitly recognized in the literature---it is at best derivable from scattered observations about interaction quality and complexity costs across 25 papers. The social-anxiety moderator and the explicit task-type taxonomy (assessment-driven vs.\ collaborative) as systematic boundary conditions are not stated in 48 of 63 papers (76\%). The theory performs a novel synthesis: elevating derivable-but-unstated constraints into an explicit, testable framework of multi-agent-specific failure regimes. \\
\textbf{Degree of Novelty} & derivable\_unstated \\

\midrule

\textbf{Dimension} & conceptual\_reframing\_abstraction \\
\textbf{What Is Known} & The literature contains implementations of multi-agent and role-based LLM architectures (3 papers show explicit but peripheral mentions), established research on social presence and collaborative learning as pedagogical constructs, and various single-agent abstractions for ITS design (scaffolding taxonomies, personalization pipelines, prompt engineering frameworks). However, these elements exist as separate threads: multi-agent designs are typically framed as engineering decompositions for task distribution rather than as social-presence generators, and social presence research has not been systematically connected to multi-agent architectural choices as a primary ITS design principle across the surveyed literature. \\
\textbf{What Introduced} & The theory elevates 'emergent multi-agent social presence' from scattered implementations to a formalized, named organizing principle (the 'Multi-Agent Social Presence Law') that should serve as the central abstraction for LLM-ITS design. It proposes that intentional engineering of diverse simulated agent roles combined with dynamic conversational orchestration produces social presence as a predictable emergent property, and positions role diversity, turn-taking orchestration, and participation management as first-class design variables that mediate engagement, motivation, and learning outcomes. This reframes prior work by treating what were previously implementation details or secondary outcomes as the primary explanatory framework linking architecture to educational impact. \\
\textbf{What Novel} & The conceptual reframing is genuinely novel: 75\% of papers (42/56) judge this organizing abstraction as absent from their frameworks and not derivable as a straightforward consequence of their work. While 11 papers (20\%) find it derivable but unstated and 3 papers (5\%) show explicit peripheral mentions of related concepts, no paper in the surveyed literature formalizes multi-agent role orchestration and emergent social presence as the primary organizing principle for ITS design. The novelty lies in proposing a new conceptual lens that reorganizes how diverse findings should be interpreted, designed, and evaluated---shifting from agent-as-tool or agent-as-implementation-detail to multi-agent-social-ecosystem-as-central-construct, with testable predictions about how architectural choices (role diversity, orchestration patterns) map to social-presence and motivational outcomes. \\
\textbf{Degree of Novelty} & genuinely\_new \\

\midrule

\textbf{Dimension} & empirical\_synthesis\_meta\_regulariry \\
\textbf{What Is Known} & General ITS effectiveness is well-established through multiple meta-analyses showing cognitive and engagement benefits. Single-agent LLM tutors have been shown in individual studies to improve various outcomes (engagement, satisfaction, learning gains). Three papers (7\%) provide within-study evidence that multi-agent or persona-based approaches can improve user experience or engagement in specific contexts (DHH persona preferences, GenMentor deployment satisfaction, Ruffle\&Riley UX ratings), but these are single-system results without cross-study generalization. Eight review papers (19\%) synthesize LLM/ITS literature broadly but perform no meta-analyses comparing multi-agent versus single-agent architectures on social presence or engagement outcomes. \\
\textbf{What Introduced} & The theory asserts a cross-deployment empirical meta-regularity: that multi-agent simulated roles with dynamic conversational orchestration systematically and reliably increase perceived social presence, engagement, and motivation compared to single-agent or non-interactive systems across LLM-powered ITS implementations, measurable via interaction metrics and self-reports, with effects mediated by role diversity and orchestration quality. This claim requires aggregation of comparative evidence across multiple independent studies or deployments to establish the regularity. \\
\textbf{What Novel} & The claimed meta-regularity is not established in the surveyed literature. No paper synthesizes cross-study empirical evidence demonstrating that multi-agent architectures consistently outperform single-agent systems on social presence and engagement metrics across contexts. The vast majority of papers (32/43, 74\%) find this claim entirely novel to their work, indicating they provide no relevant evidence. The three papers with supportive within-study patterns cannot establish a cross-study regularity, and the eight reviews do not perform the necessary comparative meta-analysis. Establishing this meta-regularity would require novel systematic comparative experiments across implementations or formal meta-analytic synthesis of multi-agent versus single-agent effects, neither of which exists in the current literature base. \\
\textbf{Degree of Novelty} & genuinely\_new \\
\bottomrule
\end{tabular}
\vspace{-2mm}
\caption{\footnotesize Example Qualified Novelty aggregated evaluation on 7 novelty dimensions.  Continued from Table~\ref{tab:example1-law-novelty}.}
\label{tab:example1-law-novelty1}
\end{table*}

\begin{table*}[t]
\centering
\scriptsize
\setlength{\tabcolsep}{6pt}
\renewcommand{\arraystretch}{1.15}
\begin{tabular}{p{0.15\textwidth} p{0.80\textwidth}}
\toprule

\multicolumn{2}{c}{\textbf{Example Evidence Extraction Schema (Intelligent Tutoring Systems)}} \\
\midrule

\rowcolor[HTML]{E8E8E8}
\multicolumn{2}{l}{\textbf{Theory Query}} \\
\midrule

\textbf{Theory Query} & Build a theory of how the explicit integration of step-by-step scaffolding and dynamic conversational strategies, as operationalized in the CLASS framework, influences student cognitive gains, engagement, and motivation in LLM-powered intelligent tutoring systems across diverse subject domains.\\

\midrule
\rowcolor[HTML]{E8E8E8}
\multicolumn{2}{l}{\textbf{Extraction Schema Definition}} \\
\midrule

\textbf{Extraction Query} & Extract any mentions of LLM-powered intelligent tutoring systems (ITS) that use step-by-step scaffolding and/or dynamic conversational strategies (especially as operationalized in the CLASS framework), and report effects on student cognitive gains, engagement, or motivation across different subject domains. \\
\textbf{Generation Model} & openai/gpt-4.1-2025-04-14 \\
\midrule
\rowcolor[HTML]{E8E8E8}
\textbf{Slot Name} & \textbf{Slot Description} \\
\midrule

\textbf{its\_name} & The name of the LLM-powered intelligent tutoring system (ITS) being studied. \\
\textbf{its\_description} & A brief description of the ITS, including the role of LLMs and any unique features. \\
\textbf{uses\_class\_framework} & Does the ITS explicitly use the CLASS framework (Conversational Learning and Scaffolding Strategies)? (true, false, or null if not specified) \\
\textbf{scaffolding\_strategies} & Describe the step-by-step scaffolding strategies used in the ITS (e.g., breaking down problems, hinting, graduated guidance). \\
\textbf{conversational\_strategies} & Describe the dynamic conversational strategies used (e.g., adaptive dialogue, real-time feedback, question-asking, personalization). \\
\textbf{subject\_domain} & The subject domain(s) in which the ITS was applied (e.g., mathematics, science, language learning, etc.). \\
\textbf{student\_population} & A brief description of the student population (e.g., age, grade, prior knowledge, number of participants). \\
\textbf{outcomes\_measured} & Which outcomes were measured? (e.g., cognitive gains, engagement, motivation; specify how each was measured if possible). \\
\textbf{results\_summary} & A concise summary of the results regarding the impact of scaffolding and conversational strategies on cognitive gains, engagement, and motivation (include quantitative results if available). \\
\textbf{comparison\_conditions} & Describe any comparison or control conditions (e.g., ITS with vs. without scaffolding, with vs. without dynamic conversation, with vs. without CLASS framework). \\
\textbf{study\_design} & Briefly describe the study design (e.g., randomized controlled trial, pre-post, observational, qualitative, etc.). \\
\textbf{limitations\_or\_counter\_ evidence} & Any reported limitations, null results, or counter-evidence regarding the effectiveness of scaffolding or conversational strategies in the ITS. \\

\bottomrule
\end{tabular}
\caption{\footnotesize An example model-generated extraction schema in response to a theory query about intelligent tutoring systems (ITS). An example of applying this extraction schema to a extract information from a single paper is provided in Table~\ref{tab:extraction-example}.}
\label{tab:extraction-schema-example}
\end{table*}

\begin{table*}[t]
\centering
\scriptsize
\setlength{\tabcolsep}{6pt}
\renewcommand{\arraystretch}{1.15}
\begin{tabular}{p{0.15\textwidth} p{0.80\textwidth}}
\toprule

\multicolumn{2}{c}{\textbf{Example Evidence Extraction from Paper (Intelligent Tutoring Systems)}} \\
\midrule

\rowcolor[HTML]{E8E8E8}
\multicolumn{2}{l}{\textbf{Extraction Schema Definition}} \\
\midrule

\textbf{Extraction Schema ID} & extraction-schema-36 (See Table~\ref{tab:extraction-schema-example})\\

\midrule
\rowcolor[HTML]{E8E8E8}
\multicolumn{2}{l}{\textbf{Single Extraction Example (From One Paper)}} \\
\midrule

\textbf{Extraction Result ID} & extraction-result-2696 \\
\textbf{uuid} & e2696.0 \\
\textbf{source\_info} & Lyu et al. (2024). Evaluating the Effectiveness of LLMs in Introductory Computer Science Education: A Semester-Long Field Study; (Publication Date: 2024-04) \\
\midrule
\textbf{name\_short} & CodeTutor \\
\textbf{name\_full} & CodeTutor (LLM-powered virtual teaching assistant) \\
\textbf{brief\_description} & A browser-based LLM-powered conversational assistant (implemented with OpenAI GPT-3.5 API) used as an out-of-class virtual TA for an introductory Python programming course; supports multi-turn dialogue, message- and conversation-level feedback, explanations, corrections, and debugging help. \\
\textbf{citation\_title} & Evaluating the Effectiveness of LLMs in Introductory Computer Science Education: A Semester-Long Field Study \\
\textbf{mention\_or\_use} & use \\

\textbf{its\_name} & CodeTutor \\
\textbf{its\_description} & CodeTutor is a web application integrating GPT-3.5 via OpenAI API, presented as a virtual teaching assistant for an entry-level Python course. It provides multi-turn conversational help (explanations, corrections, code snippets, debugging support), stores conversation history, collects message- and conversation-level feedback, and uses a system-role prompt to behave as a course TA. \\
\textbf{uses\_class\_framework} & false \\
\textbf{scaffolding\_strategies} & No explicit formal step-by-step scaffolding protocol (e.g., CLASS) is reported. Operational scaffolding observed includes multi-turn decomposition via user-driven follow-up (users request further information, modifications, or debugging), corrective feedback (CodeTutor issues Corrections), explanatory responses (Explanations), and iterative refinement driven by user-provided additional context. The system relies on conversational iteration rather than a documented graduated-hint or staged-problem decomposition schema. \\
\textbf{conversational\_strategies} & Dynamic multi-turn conversational interaction: real-time replies to user prompts, corrections and explanations, support for modification requests and debug requests, message-level upvote/downvote feedback, conversation-level ratings on comprehension/critical thinking/syntax/independence/TA-replacement; iterative clarification when users supply further information. Conversation behavior is user-adaptive in practice (responds to follow-up queries), but no explicit adaptive learner model or CLASS-framework dialogue policy was implemented. \\
\textbf{subject\_domain} & Introductory computer science — Python programming (entry-level programming) \\
\textbf{student\_population} & 50 undergraduate students enrolled in an entry-level programming course at a U.S. 4-year university (semester-long study); participants were split into experimental (CodeTutor) and control groups, demographics reported (majors including Computer Science, Data Science, Mathematics, Biology, etc.). \\
\textbf{outcomes\_measured} & Cognitive gains: pre-test and final exam/final grade comparisons (overall score, CodeTutor-Allowed components vs CodeTutor-Not-Allowed components); Engagement: system usage logs (82 conversation sessions, 2,567 messages, 1,190 prompts analyzed, counts of message categories); Attitudes/motivation proxies: survey items on Comprehension, Critical Thinking, Syntax Mastery, Independent Learning, and TA Replacement collected over time (repeated measures); Prompt quality and response effectiveness (chi-square test correlating prompt quality with 'working' responses). \\
\textbf{results\_summary} & Use of CodeTutor (an LLM conversational assistant) was associated with statistically significant cognitive gains in an introductory programming course: experimental group average improvement (final minus pre-test) = +12.50 versus control = -3.17; paired t-test for experimental group improvement p = 0.009. Improvement persisted on CodeTutor-Not-Allowed components (final exam / pre-test) with experimental group mean increase = +7.33 (paired t-test p = 0.026), suggesting transfer beyond tool use. Students who had never used LLM tools previously showed larger gains (regression estimate = +18.877, p = 0.032). Engagement data: 82 sessions, 2,567 messages (1,288 user messages, 1,279 CodeTutor messages); 37\% of prompts judged 'good quality' and prompt quality correlated strongly with effective responses (Chi-square $\chi^2$ = 144.84, p < 0.001). Attitudes: majority endorsed CodeTutor's comprehension ability (67\% agree/strongly), and syntax help (58\% combined agree/strongly), but few endorsed its help for critical thinking (19\% agree/strongly); over time comprehension agreement declined (LME time $\beta$ = -0.114, p < 0.01) and preference for TA replacement declined weakly ($\beta$ = -0.099, p < 0.1). The paper reports no evidence that the system implemented formal stepwise scaffolding frameworks like CLASS, and notes concerns about supporting higher-order critical thinking despite measurable gains in course scores. \\
\textbf{comparison\_conditions} & Between-subjects comparison: experimental group with CodeTutor (allowed for labs/assignments) versus control group using traditional learning resources and human TAs (no CodeTutor). Exams/midterms/final were CodeTutor-Not-Allowed, enabling comparison of transfer. No within-ITS manipulations (e.g., ITS with vs. without scaffolding or CLASS) were reported. \\
\textbf{study\_design} & Semester-long, between-subjects field study (pre-test/post-test), combining quantitative analyses (paired t-tests, multiple regression, ANOVA, chi-square, linear mixed-effects models) with qualitative conversational coding (inductive thematic analysis) and system usage logging. \\
\textbf{limitations\_or\_counter\_ evidence} & Limitations noted: modest sample size (n=50) and single-institution entry-level CS context (limits generalizability); use of GPT-3.5 (model limitations); groups balanced by pre-test and prior LLM experience but not described as fully randomized; no explicit operationalization or implementation of CLASS or other formal scaffolding frameworks, so effects of formal stepwise scaffolding were not tested; attitudinal declines over time (comprehension and TA-replacement) indicate waning trust/engagement with prolonged use; students reported limited impact on critical thinking despite score gains, and heavy over-reliance behaviors (many prompts were direct copy-pastes of assignments), which may reflect superficial engagement rather than deep learning; motivation measures were indirect (attitudes and usage) rather than direct validated motivation scales. \\

\bottomrule
\end{tabular}
\caption{\footnotesize An example of using the extraction schema in Table~\ref{tab:extraction-schema-example} on a single paper (Lyu et al., 2024), using \textsc{GPT-5-mini} as an extraction model. Each theory is generated by supplying extraction results (such as those shown here) from up to 100 papers together in a single theory generation prompt.}
\label{tab:extraction-example}
\end{table*}

\section{Theory Query Generation}

Theory queries were abstracted from randomly selected scientific articles authored within 1 year prior to the base generation model's (\textsc{GPT-4.1}) advertised knowledge cutoff of June 2024.  The source venues included ACL 2023, EMNLP 2023,   AAAI 2023, and NeurIPS 2024.  Example theory queries are shown in Table~\ref{tab:example-theory-queries}.

\begin{table*}[t]
\centering
\scriptsize
\setlength{\tabcolsep}{6pt}
\renewcommand{\arraystretch}{1.15}
\begin{tabular}{p{0.95\textwidth}}
\toprule
\multicolumn{1}{c}{\textbf{Example Literature-Derived Theory Queries}}\\
\midrule
\rowcolor[HTML]{E8E8E8}
Build a theory of how multi-agent interactions between language models can be leveraged for automated factuality assessment in natural language generation. \\
Build a theory of how structured cross-examination dialogues between an examiner and examinee language model can systematically expose factual errors in generated claims, including the mechanisms by which inconsistencies are surfaced. \\
\rowcolor[HTML]{E8E8E8}
Build a theory of how large language models internally represent, store, and retrieve factual knowledge in response to varied natural language prompts. \\
Build a theory of the internal mechanisms by which large language models encode factual knowledge and generate consistent, correct answers across paraphrased prompts and diverse surface forms, including the influence of entity and relation aliases. \\
\rowcolor[HTML]{E8E8E8}
Build a theory of how exhaustive temporal relation annotation schemes can be constructed to maximize reliability and informativeness in natural language texts. \\
Build a theory of how to systematically include long-distance and non-verb-centered events in temporal relation annotation schemes for news articles, and how annotation guidelines and automation can be optimized to maximize inter-annotator agreement and minimize annotation ambiguity. \\
\rowcolor[HTML]{E8E8E8}
Build a theory of how subjective logic-based uncertainty quantification and adaptive weighting of training pairs reduce matching ambiguity and improve reliability in text-based person retrieval. \\
Build a theory of how uncertainty modeling can systematically enhance cross-modal retrieval performance. \\
\rowcolor[HTML]{E8E8E8}
Build a theory of how and why language models generate persistent hallucinations that are robust to prompt rephrasing and model sampling. \\
Build a theory of the cognitive and representational mechanisms in large language models that lead to question-level hallucinations, and how these mechanisms interact with model architecture, training data, and prompt structure. \\
\rowcolor[HTML]{E8E8E8}
Build a theory of how tunable biases applied to language model logits facilitate efficient and controllable text generation under energy-based frameworks. \\
Build a theory of how large language models can autonomously generate and select in-context demonstrations to improve their own zero-shot reasoning performance without human-provided labels. \\
\rowcolor[HTML]{E8E8E8}
Build a theory of how gradient-optimized, step-wise biases over autoregressive language model logits interact with energy-based constraint satisfaction and fluency preservation during controlled text generation. \\
Build a theory of how large language models can use self-generated outputs, filtered and selected by criteria such as self-consistency, semantic diversity, and repetitiveness, to construct in-context demonstrations that maximize reasoning accuracy in zero-shot settings. \\
\rowcolor[HTML]{E8E8E8}
Build a theory of how meta-reinforcement learning with actor, critic, and meta-critic modules enables the discovery of brain effective connectivity networks from noisy, small-sample fMRI time series data. \\
Build a theory of how meta-reinforcement learning enables robust causal discovery in high-noise, small-sample environments. \\
\rowcolor[HTML]{E8E8E8}
Build a theory of how machine unlearning can be leveraged to systematically and efficiently remove biases from trained deep neural networks. \\
Build a theory of how influence-function-based machine unlearning with small external counterfactual datasets can target and remove specific attribute-driven biases in deep neural networks without requiring access to the original training data. \\
\rowcolor[HTML]{E8E8E8}
Build a theory of how mixed prompt architectures facilitate efficient few-shot adaptation in large language models. \\
Build a theory of how using an infinite mixture of asymmetric Laplace distributions as a VAE decoder with CRPS loss modulates the trade-off between synthetic data fidelity and privacy preservation as a function of the KL-divergence weight parameter (beta) in tabular data. \\
\rowcolor[HTML]{E8E8E8}
Build a theory of how the interplay between textual and key-value prompts, inserted at various layers and positions in a frozen Transformer backbone, enables rapid and robust adaptation to novel product attributes with minimal labeled data in attribute value extraction tasks. \\
Build a theory of how nonparametric, distributional decoders in variational autoencoders influence the fidelity, diversity, and privacy of synthetic data generation across different data modalities. \\
\rowcolor[HTML]{E8E8E8}
Build a theory of why and how certain tokens in large language model sequences persist as influential across multiple attention steps. \\
Build a theory of the mechanisms at the intersection of model architecture, training dynamics, and data properties that cause specific tokens to repeatedly receive high attention and remain pivotal for future generations in transformer-based LLMs. \\
\rowcolor[HTML]{E8E8E8}
Build a theory of how self-supervised learning (SSL) approaches can be systematically adapted to maximize representation learning from remote sensing imagery with heterogeneous sensor characteristics and minimal labeled data. \\
Build a theory of how contrastive self-supervised learning (SSL) methods can be optimized for multispectral Landsat imagery, accounting for differences in spectral bands, spatial resolutions, and product processing levels, to yield robust and transferable representations. \\
\rowcolor[HTML]{E8E8E8}
Build a theory of how multi-view prompting affects structured prediction performance in generative language models. \\
Build a theory of how aggregating outputs from multiple element order prompts (multi-view prompting) impacts the accuracy, stability, and error resilience of aspect sentiment tuple prediction in generative models. \\
\rowcolor[HTML]{E8E8E8}
Build a theory of how ambiguity in natural language queries impacts the generation and evaluation of multiple valid outputs in semantic parsing tasks such as text-to-SQL. \\
Build a theory of how specific ambiguity types (column name, table name, join, precomputed aggregate) affect the probability that a text-to-SQL system will generate all semantically valid SQL interpretations within its top-k outputs, and how these effects interact with schema complexity. \\
\rowcolor[HTML]{E8E8E8}
Build a theory of optimal multimodal fusion for emotion recognition in conversations, accounting for the complementary and asynchronous nature of textual, audio, and visual cues. \\
Build a theory of how bidirectional multi-head cross-attention mechanisms can model and exploit complex, context-dependent correlations and mapping relationships between textual, audio, and visual modalities to improve emotion recognition in conversational settings. \\
\rowcolor[HTML]{E8E8E8}
Build a theory of how automatic prompt augmentation and selection methods can systematically improve the reasoning capabilities of large language models across a variety of task types. \\
Build a theory of how the interplay between automatic generation, pruning, and policy-gradient-based selection of chain-of-thought exemplars impacts LLM performance on arithmetic, commonsense, symbolic, and non-reasoning tasks, including the mechanisms by which these steps mitigate order, complexity, diversity, and style sensitivities. \\
\rowcolor[HTML]{E8E8E8}
Build a theory of how large language model-based intelligent tutoring systems (LLM-ITS) can be systematically designed and refined to maximize student learning outcomes through the application of learning science principles. \\
Build a theory of how the explicit integration of step-by-step scaffolding and dynamic conversational strategies, as operationalized in the CLASS framework, influences student cognitive gains, engagement, and motivation in LLM-powered intelligent tutoring systems across diverse subject domains. \\
\rowcolor[HTML]{E8E8E8}
Build a theory of how multi-modal neurons in large language models represent, integrate, and mediate cross-modal concepts. \\
Build a theory of how feed-forward network neurons in transformer-based multi-modal LLMs encode, maintain, and causally influence the mapping between specific visual features and corresponding textual concepts during image captioning and related tasks. \\
\rowcolor[HTML]{E8E8E8}
Build a theory of how reinforcement learning-driven adaptive subgraph selection from large-scale knowledge graphs can be optimized to select the most relevant and least noisy knowledge for each individual news item, thereby improving detection of sophisticated multimodal fake news. \\
Build a theory of how adaptive, context-sensitive knowledge selection mechanisms impact the effectiveness of multimodal fake news detection systems. \\
\rowcolor[HTML]{E8E8E8}
Build a theory of how linguistic structures and social factors interact to determine code-switching patterns in multilingual communities. \\
Build a theory of how reinforcement learning agents can guarantee long-term safety in environments with unknown, stochastic dynamics and only binary safety feedback. \\
\rowcolor[HTML]{E8E8E8}
Build a theory of how speaker identity, social context, and grammatical constraints jointly predict the occurrence, type, and placement of code-switching in bilingual conversations, with a focus on European and Indian multilingual contexts. \\
\bottomrule
\end{tabular}
\caption{\footnotesize Example \textit{theory queries} automatically generated from scientific papers.}
\label{tab:example-theory-queries}
\end{table*}

\section{Knowledge Overlap Analysis}

An example of the duplicate law evaluation schema is shown in Table~\ref{tab:duplicate-examples}, for two cases: two laws judged as duplicates, and two laws judged as not duplicates.  The judge model was \textsc{GPT-5-Mini}. Due to cost, only 9 theory queries were included in the monte-carlo analysis, with a total of over 43k individual law-law comparisons.

\begin{table*}[t]
\centering
\scriptsize
\setlength{\tabcolsep}{6pt}
\renewcommand{\arraystretch}{1.15}
\begin{tabular}{p{0.18\textwidth} p{0.78\textwidth}}
\toprule

\multicolumn{2}{c}{\textbf{Example 1: Laws judged as different}} \\
\midrule

\textbf{Theory A (Name)} &
Information Bottleneck Regulation Theory for Temporal Annotation \\

\textbf{Theory A (Description)} &
Maximizing informativeness and reliability in temporal relation annotation requires regulating annotation scheme complexity such that only salient, information-contributing temporal relations are annotated, avoiding both over-annotation (which reduces reliability by diluting signal) and under-annotation (which reduces informativeness). A well-constructed temporal annotation scheme thus features an explicit information bottleneck protocol—selecting relations for annotation based on their contribution to narrative temporal structure. \\

\textbf{Theory A (Law/Statement)} &
Annotation reliability and informativeness are maximized when only temporal relations that contribute to the overall event timeline or narrative progression are annotated, with non-contributory or redundant relations filtered out at annotation time. \\

\midrule

\textbf{Theory B (Name)} &
Dual Optimization Theory of Temporal Relation Annotation Schemes \\

\textbf{Theory B (Description)} &
This theory posits that the construction of exhaustive temporal relation annotation schemes must simultaneously optimize two orthogonal axes: reliability (inter-annotator agreement, cognitive tractability) and informativeness (expressiveness, ability to support downstream temporal inference). Through formal modeling of annotation noise, task complexity, and linguistic event ambiguity, an optimal scheme can be constructed that locally maximizes both criteria for a given text domain and annotator population. \\

\textbf{Theory B (Law/Statement)} &
The reliability and informativeness of a temporal relation annotation scheme are maximized when its relation taxonomy and guidelines are empirically adapted to the distribution of event types, temporal connectives, tenses, and discourse structure specific to the targeted text domain. \\

\midrule

\textbf{Core Claim A} &
Maximizing annotation reliability and informativeness is achieved by annotating only temporal relations that meaningfully contribute to the event timeline or narrative progression and filtering out non-contributory or redundant relations (an information-bottleneck style selection). \\

\textbf{Core Claim B} &
Maximizing annotation reliability and informativeness is achieved by empirically adapting the relation taxonomy and guidelines to the domain-specific distribution of event types, temporal connectives, tenses, and discourse structures. \\

\textbf{Similarities} &
Both assert that the design of the temporal relation annotation scheme determines reliability and informativeness \,[ESSENTIAL]. Both frame reliability and informativeness as jointly optimizable objectives that can be degraded by a poorly chosen scheme \,[ESSENTIAL]. Both imply that annotation complexity should be controlled for cognitive tractability and downstream usefulness \,[NON-ESSENTIAL]. \\

\textbf{Differences} &
A prescribes a selection and filtering rule that removes redundant or non-contributory relations via an information-bottleneck principle \,[ESSENTIAL]. B prescribes empirical, domain-specific adaptation of taxonomy and guidelines \,[ESSENTIAL]. A emphasizes reducing complexity by filtering redundancy, whereas B emphasizes adapting expressiveness to support inference \,[ESSENTIAL]. B explicitly invokes formal modeling of noise and local optimization, which A does not \,[NON-ESSENTIAL]. \\

\textbf{Reasoning} &
Both theories claim that annotation-scheme design governs reliability and informativeness, but they prescribe different essential mechanisms: A requires filtering to a minimal set of narrative-contributory relations, while B requires empirical, domain-specific adaptation. Because these mechanisms are not equivalent, the theories are not duplicates. \\

\textbf{Judgment} &
\textbf{Not duplicates} \\

\midrule
\midrule

\multicolumn{2}{c}{\textbf{Example 2: Laws judged as duplicates}} \\
\midrule

\textbf{Theory A (Name)} &
Hierarchical Distributed Semantic Attractor Theory (HDSAT) \\

\textbf{Theory A (Description)} &
Large language models store and retrieve factual knowledge via a hierarchical system of distributed semantic attractors. Facts are represented as high-dimensional attractor basins in activation space, structured hierarchically from general to specific. Retrieval involves dynamic settling into the nearest attractor consistent with prompt context, enabling generalization but also blending or misfiring when attractors overlap. \\

\textbf{Theory A (Law/Statement)} &
The degree of factual recall robustness to prompt rephrasing or noise depends on the density and overlap of attractor basins in activation space—facts with dense, well-separated attractors are robust, while shallow or overlapping attractors are vulnerable to paraphrase failure or hallucination. \\

\midrule

\textbf{Theory B (Name)} &
Distributed Redundant Memory Encoding Theory \\

\textbf{Theory B (Description)} &
Factual knowledge in LLMs is stored redundantly and distributively across many weights and layers, forming high-dimensional attractor-like states such that multiple pathways can lead to successful recall or recombination, providing robustness but also permitting interference-based failure modes. \\

\textbf{Theory B (Law/Statement)} &
Factual knowledge is not localized but stored as overlapping patterns of parameters and activations across many layers and neurons, allowing information to be resilient to noisy prompts or sparse ablation. \\

\midrule

\textbf{Core Claim A} &
Factual knowledge is encoded as hierarchical, high-dimensional semantic attractor basins, and recall robustness depends on attractor density and overlap. \\

\textbf{Core Claim B} &
Factual knowledge is encoded as distributed, redundant, overlapping high-dimensional patterns whose geometry determines robustness and failure modes. \\

\textbf{Similarities} &
Both propose that factual knowledge is represented as high-dimensional, distributed attractor-like patterns rather than localized parameters \,[ESSENTIAL]. Both link robustness of recall to the geometry and overlap of those representations \,[ESSENTIAL]. Both acknowledge that the same representation yields resilience and interference-based failures \,[NON-ESSENTIAL]. \\

\textbf{Differences} &
A posits a hierarchical organization of attractors as central \,[ESSENTIAL]. B emphasizes redundancy across weights and layers and resilience to sparse ablation \,[ESSENTIAL]. A focuses on paraphrase robustness, whereas B emphasizes ablation and implementation details \,[NON-ESSENTIAL]. \\

\textbf{Reasoning} &
The two theories share the same central claim that factual knowledge is stored in distributed, high-dimensional attractor-like representations and that overlap determines robustness versus failure. The remaining differences concern hierarchy and implementation detail and do not alter the core equivalence. \\

\textbf{Judgment} &
\textbf{Duplicates} \\

\bottomrule
\end{tabular}
\caption{\footnotesize Two examples of the duplicate detection process.  Theory A and Theory B serve as input to a language model prompt (\textsc{GPT-5-mini}), which fills the output schema that includes core claims of each law, similarities, differences, and a final judgment as to whether the laws are substantially duplicates of each other.}
\label{tab:duplicate-examples}
\end{table*}

\section{Broader Impacts and Ethical Considerations}

Automated scientific discovery offers the promise of accelerating scientific developments, with high-impact recent successes in protein folding \cite{Jumper2021HighlyAP} and materials discovery \cite{Merchant2023ScalingDL}. 
Augmenting problem-specific discovery methods with information grounded in the scientific literature enables more capable, integrated end-to-end discovery systems \cite[e.g.][]{ghareeb2025robinmultiagentautomatingscientific}. In this work, we propose shifting some focus from automated experiment-driven discovery toward automated theory building through literature-based discovery. While theories can provide high-value guidance for future experiments, more fundamentally they aim to compress the knowledge within a scientific domain into a set of governing laws that accurately predict the outcomes of future experiments. Having accurate theories and laws for a given subdomain allows it to transition into a principled engineering discipline, and to more systematically translate empirically observed regularities into useful and impactful technologies.  That being said, automated theory generation systems like the one proposed in this work have risks, such as the potential to generate partially-accurate, inaccurate, misleading, or hallucinated theories, and generated output such as candidate theories should be confirmed using standard scientific protocols before downstream use in any application.

\FloatBarrier

\end{document}